\documentclass{article}
\usepackage{graphicx}

\usepackage{algorithmic}
\usepackage{algorithm}
\usepackage[caption=false,font=normalsize,labelfont=sf,textfont=sf]{subfig}
\usepackage{textcomp}
\usepackage{stfloats}
\usepackage{url}
\usepackage{verbatim}
\usepackage{cite}
\usepackage{cuted}
\usepackage{caption}
\usepackage{pifont}
\usepackage{array} 
\usepackage{subcaption} 

\usepackage{multirow}

\usepackage{microtype}
\usepackage{booktabs} 

\usepackage{hyperref}

\usepackage[accepted]{icml2025}

\usepackage{amsmath}
\usepackage{amssymb}
\usepackage{mathtools}
\usepackage{amsthm}

\newcommand*{\method}{TripoSG}

\usepackage[textsize=tiny]{todonotes}

\icmltitlerunning{\method{}: High-Fidelity 3D Shape Synthesis using Large-Scale Rectified Flow Models}

\begin{document}

\twocolumn[
\icmltitle{\method{}: High-Fidelity 3D Shape Synthesis using Large-Scale Rectified Flow Models}

\icmlsetsymbol{equal}{*}
\icmlsetsymbol{corre}{$\dagger$}

\begin{icmlauthorlist}
\icmlauthor{Yangguang Li$^1$}{equal}
\icmlauthor{Zi-Xin Zou$^1$}{equal}
\icmlauthor{Zexiang Liu$^1$}{}
\icmlauthor{Dehu Wang$^1$}{}
\icmlauthor{Yuan Liang$^1$}{}
\icmlauthor{Zhipeng Yu$^1$}{}
\icmlauthor{Xingchao Liu$^3$}{}
\icmlauthor{Yuan-Chen Guo$^1$}{}
\icmlauthor{Ding Liang$^1$}{corre}
\icmlauthor{Wanli Ouyang$^{2,4}$}{}
\icmlauthor{Yan-Pei Cao$^1$}{}\\
~\\
Project Homepage: \href{https://yg256li.github.io/TripoSG-Page/}{\textcolor{red}{TripoSG-Page}}
\end{icmlauthorlist}

\icmlkeywords{Machine Learning, ICML}

\vskip 0.3in
]




\printAffiliationsAndNotice
{
~\\
$^1$VAST \\
$^2$The Chinese University of Hong Kong \\
$^3$The University of Texas at Austin \\
$^4$Shanghai AI Laboratory \\
$^*$: Equal Contribution \\
$^\dagger$: Corresponding Author\\
}

\begin{strip}
    \centering
    \vspace{-9em}
    \includegraphics[width=1\textwidth]{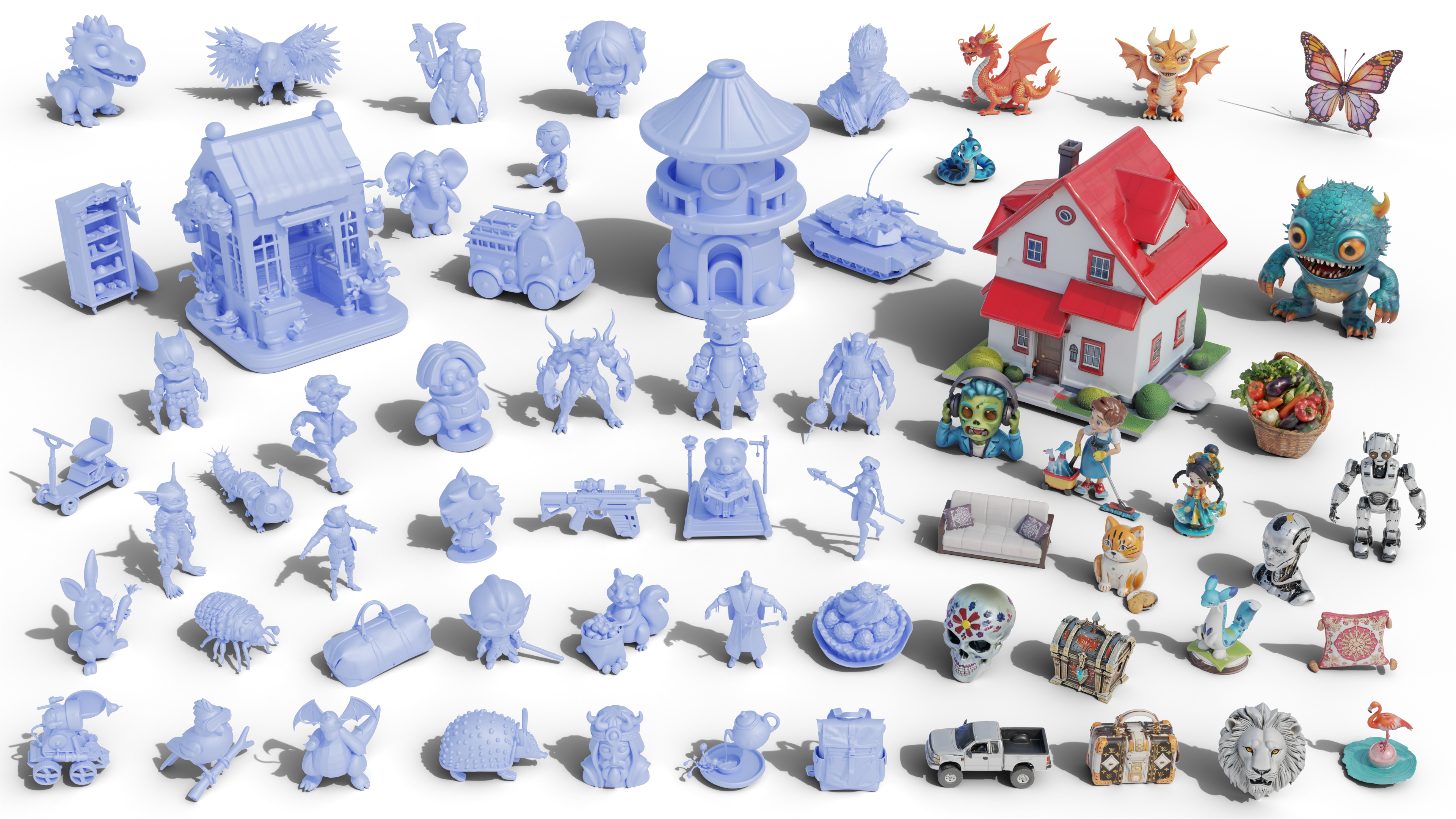}
    \captionof{figure}{High-quality 3D shape samples from our largest \method{} model. Covering various complex structures, diverse styles, imaginative designs, multi-object compositions, and richly detailed outputs, demonstrates its powerful generation capabilities.}
    \vspace{2em}
    \label{fig:teaser}
\end{strip}

\begin{abstract}
Recent advancements in diffusion techniques have propelled image and video generation to unprecedented levels of quality, significantly accelerating the deployment and application of generative AI. However, 3D shape generation technology has so far lagged behind, constrained by limitations in 3D data scale, complexity of 3D data processing, and insufficient exploration of advanced techniques in the 3D domain.
Current approaches to 3D shape generation face substantial challenges in terms of output quality, generalization capability, and alignment with input conditions.
We present \method{}, a new streamlined shape diffusion paradigm capable of generating high-fidelity 3D meshes with precise correspondence to input images.
Specifically, we propose: 
1) A large-scale rectified flow transformer for 3D shape generation, achieving state-of-the-art fidelity through training on extensive, high-quality data.
2) A hybrid supervised training strategy combining SDF, normal, and eikonal losses for 3D VAE, achieving high-quality 3D reconstruction performance.
3) A data processing pipeline to generate 2 million high-quality 3D samples, highlighting the crucial rules for data quality and quantity in training 3D generative models.
Through comprehensive experiments, we have validated the effectiveness of each component in our new framework. The seamless integration of these parts has enabled \method{} to achieve state-of-the-art performance in 3D shape generation.
The resulting 3D shapes exhibit enhanced detail due to high-resolution capabilities and demonstrate exceptional fidelity to input images.
Moreover, \method{} demonstrates improved versatility in generating 3D models from diverse image styles and contents, showcasing strong generalization capabilities.
To foster progress and innovation in the field of 3D generation, we open source our model and code at: \href{https://github.com/VAST-AI-Research/TripoSG}{\textcolor{red}{TripoSG}}.
\end{abstract}

\section{Introduction}
Recent advancements in large-scale visual datasets \cite{schuhmann2022laion, bain2021frozen, wang2023internvid} have propelled remarkable progress in generative models. These models effectively compress high-dimensional visual data, such as images and videos, into latent spaces, enabling the generation of high-quality visual content conditioned on various input modalities. State-of-the-art generative AI models, including SD3~\cite{esser2024scaling}, FLUX~\cite{flux}, and Sora~\cite{videoworldsimulators2024}, exemplify this capability, producing strikingly realistic images and videos from diverse conditional inputs. This breakthrough has revolutionized human visual creation, opening new avenues for artistic expression and content generation.

In the domain of 3D content creation, the pursuit of high-quality, production-ready 3D generation remains a primary objective for researchers, artists, and designers. Substantial progress has been made in generating 3D models from single images, with approaches broadly categorized into two paradigms: large-scale reconstruction-based methods~\cite{hong2023lrm,li2023instant3d,wang2023pf,xu2023dmv3d,zhang2024gs,wei2024meshlrm,xu2024instantmesh,wang2024crm,tochilkin2024triposr,zou2024triplane} and diffusion-based methods~\cite{zhang20233dshape2vecset,zhao2024michelangelo,zhang2024clay,wu2024direct3d,li2024craftsman}.
Large-scale reconstruction-based methods primarily utilize a network to regress the 3D model in a deterministic way. While effective, these approaches often struggle with inconsistencies in overlapping regions from multiple input views (which can be generated from a single view by multi-view diffusion models~\cite{shi2023mvdream,wang2023imagedream,liu2023unidream}) and exhibit artifacts in occluded areas.
Conversely, diffusion-based methods train on 3D representations or latent representations compressed by Variational AutoEncoders (VAEs). As generative rather than regression methods, they circumvent some challenges inherent to reconstruction approaches. However, current methods predominantly rely on occupancy representations, often necessitating additional post-processing to mitigate aliasing artifacts and lacking fine-grained geometric details. Moreover, vanilla diffusion architectures and sampling strategies yield suboptimal 3D model quality, resulting in a significant alignment gap between generated models and input images.
A common limitation across both approaches is their heavy reliance on the Objaverse dataset \cite{deitke2023objaverse}. The necessity for rigorous data filtering often reduces the usable data samples by nearly half, presenting a substantial challenge in scaling data compared to the image and video domains.

Given these challenges in 3D generation and the substantial successes observed in image and video synthesis, we posit a critical question: \emph{What is the optimal paradigm for generating high-fidelity 3D models with precise alignment to input conditions?}

\begin{figure}
    \centering
    \includegraphics[width=\linewidth]{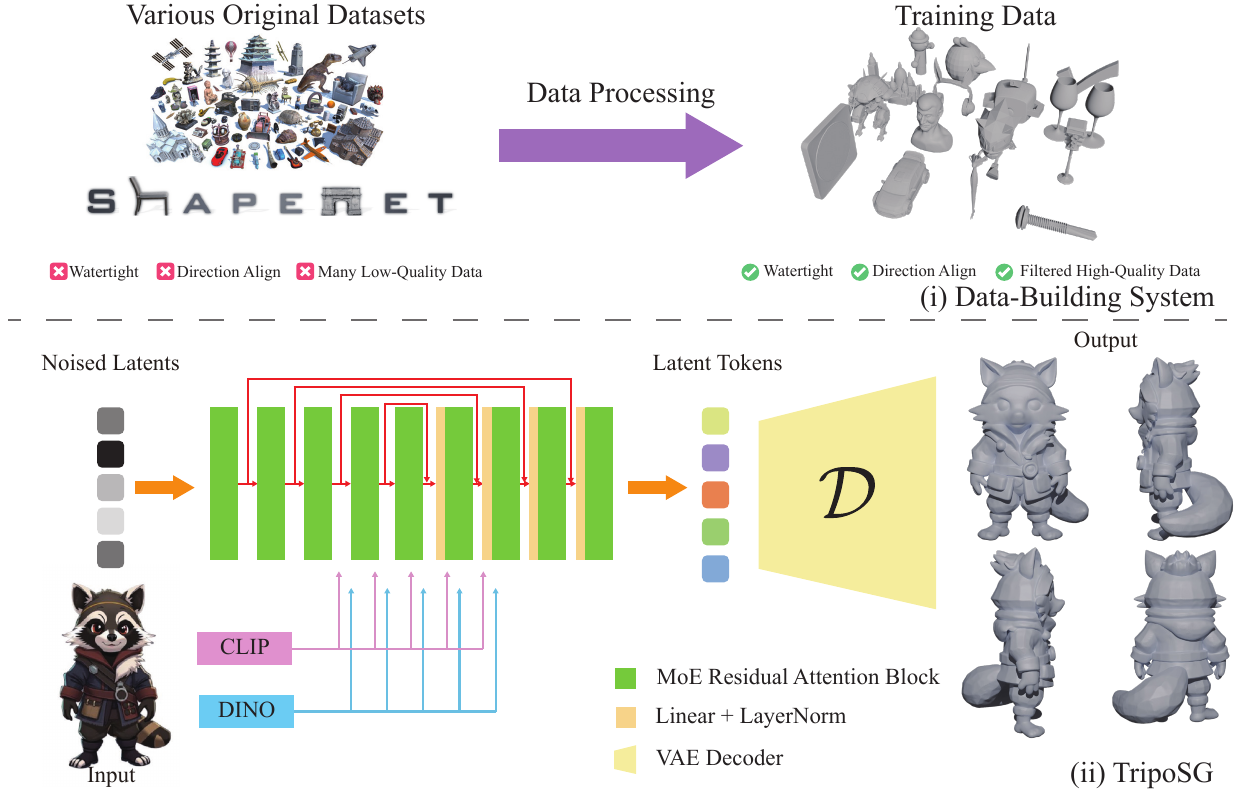}
    \caption{The overview of our method consists of two main components: (i) Data-Building System and (ii) \method{} Model. The data-building system processes the 3D models from various datasets (e.g., Objaverse and ShapeNet) through a series of data processing steps to create the training data. Our \method{} model is then trained on this curated dataset for high-fidelity shape generation from a single input image.}
    \label{fig:pipeline}
\end{figure}

In response to this inquiry, we present \emph{\method{}}, a high-fidelity 3D generative model that leverages a rectified flow transformer trained on meticulously curated, large-scale data. Our approach achieves unprecedented quality in image-to-3D generation, characterized by finer details and superior input condition alignment.
Inspired by 3DShape2VecSet~\cite{zhang20233dshape2vecset}, we train our generative model on latent representations efficiently compressed by a VAE model.
We identify that the quality of the generated models is heavily dependent on the capacity of the latent space (i.e., the number of tokens) and the volume of high-quality 3D data available.
Thus, it's crucial to scale up the model with a more efficient neural architecture. To address this challenge, we propose several key advancements:
\begin{enumerate}
\item We pioneer the use of a rectified flow transformer architecture~\cite{DBLP:conf/iclr/LiuG023} in 3D generation, drawing inspiration from recent successes in scaling up image/video models (e.g., SD3~\cite{esser2024scaling}, FLUX
~\cite{flux}, and Meta Movie Gen~\cite{metamoviegen}). This choice is motivated by its simplicity and superior performance in terms of training stability and convergence.
\item Our model builds upon the DiT~\cite{peebles2023scalable} framework, incorporating critical enhancements for improved scalability. These include skip-connections, RMSNorm~\cite{zhang2019root}, and the injection of both global and local features. Furthermore, we adopt the DiT-MoE~\cite{fei2024scaling} approach, replacing standard Feed-Forward modules in each block with a mixture-of-experts (MoE) mechanism. This configuration retains one shared expert while selecting the top two experts for each operation.
\item We have successfully trained a 4 billion (4B) parameter 3D generative model at a latent resolution of $4096$ tokens, leveraging this unprecedented scale to produce highly detailed and geometrically precise structures.
\end{enumerate}

Recognizing the critical role of VAE reconstruction quality in latent diffusion or flow models~\cite{rombach2022high,zhang20233dshape2vecset}, we introduce several improvements to our VAE training process, including \textbf{1)} We employ SDF (Signed Distance Function) representation, which offers superior geometric expressiveness compared to occupancy representations. Occupancy grids often introduce quantization errors, resulting in aliasing artifacts or ``staircasing'' effects when representing continuous surfaces. \textbf{2)} We implement geometry-aware supervision on SDF with surface normal guidance, significantly enhancing 3D model reconstruction. This approach enables sharper and more precise geometry without the quantization artifacts.
These improvements can better connect the latent space and the 3D model space, benefiting the quality of 3D models generated by our flow transformers.

To address the scarcity of high-quality 3D data, we have developed a sophisticated data-building system. This system ensures the generation of standardized, high-quality 3D training data (Image-SDF pairs) from diverse sources through a rigorous process of 1) scoring, 2) filtering, 3) fixing and augmentation, and 4) field data production.
More importantly, we find that \emph{both} data quality and quantity are crucial, demonstrating that improperly processed data can substantially impede the training process.

Building on our proposed solution, we have designed an optimized training configuration that incorporates the proposed improvements, achieving a new state-of-the-art (SOTA) performance in 3D generation with our largest \method{} model.
Fig.~\ref{fig:pipeline} provides an overview of our method, illustrating both the data-building system and the \method{} model architecture. 
We have validated the effectiveness of each design component through a series of mini-setting experiments.

Our core contributions are as follows:

\begin{itemize}
    \item We introduce a large-scale rectified flow transformer for 3D shape generation, setting a new benchmark in fidelity by leveraging extensive, high-quality data.
    \item We present a novel hybrid supervised training strategy that combines SDF, normal, and eikonal losses for 3D VAE, achieving state-of-the-art 3D reconstruction performance.
    \item We demonstrate the crucial rule of data quality and quantity in training 3D generative models, introducing a robust data processing pipeline capable of producing 2M high-quality 3D data samples.
\end{itemize}
\section{Related Work} 
\subsection{Lifting 2D Prior to 3D Modeling}
The diffusion model~\cite{DBLP:conf/nips/HoJA20} has demonstrated strong generative capabilities in image~\cite{rombach2022high,DBLP:conf/icml/RameshPGGVRCS21} or video~\cite{DBLP:conf/iclr/SingerPH00ZHYAG23,DBLP:conf/iccv/WuGWLGSHSQS23} generation. However, due to the limitations of high-quality 3D data, it has long been challenging to directly transfer the techniques from text and image generation to 3D tasks.
DreamFusion~\cite{DBLP:conf/iclr/PooleJBM23} pioneered the use of image diffusion priors for 3D generation by proposing a Score Distillation Sampling method enabling iterative optimization of the 3D representation via differentiable volume rendering~\cite{DBLP:conf/eccv/MildenhallSTBRN20}.
Subsequent work introduced numerous improvements in areas such as 3D represetation~\cite{DBLP:conf/cvpr/Lin0TTZHKF0L23,yi2024gaussiandreamer,DBLP:conf/iclr/TangRZ0Z24}, sampling strategy~\cite{DBLP:conf/nips/Wang00BL0023,DBLP:conf/cvpr/WangDLYS23,liang2024luciddreamer,DBLP:conf/aaai/Zou0CHSZ24}, incorporating additional geometric cues~\cite{DBLP:conf/iccv/TangWZZYM023,long2024wonder3d}, and multi-view image generation consistency~\cite{DBLP:conf/iclr/ShiWYMLY24,DBLP:conf/iclr/LiuLZLLKW24,DBLP:journals/corr/abs-2312-02201}.
Different from text-to-3D generation methods using off-the-shell text-to-image diffusion model, e.g., Stable Diffusion~\cite{rombach2022high}, many works explore to train a viewpoint-aware image diffusion based on input images~\cite{DBLP:conf/iccv/LiuWHTZV23,DBLP:conf/iccv/ChanNCBPLAMKW23,DBLP:journals/corr/abs-2310-15110}.
When image generation across multiple views becomes more consistent or with normal or depth generation, the 3D model can be directly optimized by pixel-level loss instead of time-consuming distillation sampling, resulting in 3D generation in a few minutes~\cite{long2024wonder3d,DBLP:journals/corr/abs-2405-20343,DBLP:journals/corr/abs-2405-11616}.

\subsection{Large 3D Reconstruction Modeling}
Unlike the previously introduced methods, which require a time-consuming optimization process lasting several minutes or even hours, various works propose to learn geometry with diverse representation types (e.g., point cloud~\cite{DBLP:conf/cvpr/FanSG17,DBLP:conf/cvpr/WuZXZC20}, voxel~\cite{DBLP:conf/eccv/GirdharFRG16,DBLP:conf/nips/0001WXSFT17}, mesh~\cite{DBLP:conf/eccv/WangZLFLJ18,DBLP:conf/cvpr/WorchelDHSFE22} or implicit field~\cite{DBLP:conf/cvpr/MeschederONNG19,DBLP:conf/nips/XuWCMN19,DBLP:conf/cvpr/YuYTK21}) from input images in a deterministic process, with encoder-decoder network architecture.
Recently, equipped with a ginormous collection of high-quality 3D models in Objaverse (-XL)~\cite{deitke2023objaverse,deitke2024objaverse} as well as an advanced and scalable Transformer-based architecture~\cite{DBLP:conf/nips/VaswaniSPUJGKP17}, Large Reconstruction Model (LRM)~\cite{hong2023lrm} and its many subsequent variants~\cite{xu2023dmv3d,wang2023pf,li2023instant3d,zhang2024gs,zou2024triplane,wei2024meshlrm,xu2024instantmesh,tochilkin2024triposr} has greatly promoted the development of reconstruction-based methods.
MeshFormer~\cite{DBLP:journals/corr/abs-2408-10198} additionally leverages sparse UNet to downsample the voxel for Transformer layers, leading to impressive reconstruction quality.
One-2-3-45~\cite{DBLP:conf/nips/LiuXJCTXS23} first proposes combining a 2D image diffusion model and a multiview reconstruction model, achieving generation capabilities while maintaining fast reconstruction speed.
Bridged with the text-to-image and image-to-multi-view diffusion models, these multiview reconstruction methods can be easily extended to text-to-3D or image-to-3D generation tasks and achieve impressive results.
However, the inconsistency between input images from different views can lead to a decline in reconstruction quality, and the unobserved regions may yield blurred results.
Thus, these are only `reconstruction' methods rather than `generation' methods, fundamentally limiting the quality ceiling of such methods.

\subsection{3D Diffusion Modeling}
Training a 3D diffusion model for 3D generation is a natural idea that stems from advancements in the field of image~\cite{rombach2022high,DBLP:conf/icml/RameshPGGVRCS21} and video~\cite{DBLP:conf/iclr/SingerPH00ZHYAG23,DBLP:conf/iccv/WuGWLGSHSQS23} generation.
Many previous works train a diffusion model based on various 3D representations, such as voxel~\cite{DBLP:conf/cvpr/MullerSPBKN23}, point cloud~\cite{DBLP:conf/iccv/ZhouD021,DBLP:conf/nips/zengVWGLFK22,DBLP:conf/cvpr/Melas-Kyriazi0V23}, triplane~\cite{DBLP:conf/cvpr/ShueCPA0W23}, or Occupancy/SDF grid~\cite{DBLP:journals/tog/ZhengPWTLS23,DBLP:conf/siggrapha/HuiLHF22}.
Some other works utilize a VAE to transfer the original representation to a compact latent space, and then train a diffusion model on the latent space~\cite{zhang20233dshape2vecset,DBLP:conf/cvpr/ChengLTSG23}.
For a long time, these 3D diffusion methods have struggled to match the performance of the two major categories of approaches mentioned above due to the lack of large and high-quality 3D model datasets.
These methods are mostly trained on simple 3D datasets (e.g., ShapeNet~\cite{chang2015shapenet}), with limited generation ability and effectiveness, which hinders their practical application.
Recently, some researchers have attempted to train a latent 3D diffusion model based on a large amount of high-quality 3D models~\cite{zhang2024clay,wu2024direct3d,li2024craftsman,lan2024ln3diff,DBLP:journals/corr/abs-2403-02234} and have demonstrated impressive 3D generation results.
However, these methods still have limitations on \emph{high-fidelity} generation with \emph{image alignment}.
In this paper, we adopt a 3D representation with better geometry expression ability and improve the diffusion model architecture and training strategy, achieving state-of-the-art performance on 3D shape generation.

\section{\method{}}
This section outlines the specific framework of the \method{} paradigm, which consists of three main parts: the flow-based generation architecture and sampling schedule ( Sec.\ref{sec:diffusion}); the scaling-up strategy (Sec.\ref{sec:scale_up}); the VAE architecture and supervision (Sec.\ref{sec:vae}).

\subsection{Rectified Flow Transformer}\label{sec:diffusion}

Leveraging a meticulously designed VAE architecture and robust supervision information, \method{}’s VAE, described in detail in Sec.\ref{sec:vae}, following extensive training on large-scale datasets, is capable of encoding arbitrary 3D shapes into multi-scale latent representations $X = L\times C$, $L\in \{512, 2048\}$, $C=64$, as well as decoding them back into 3D meshes. Drawing inspiration from models such as LDM~\cite{rombach2022high} and 3DShape2VecSet~\cite{zhang20233dshape2vecset}, we further train a rectified flow model on these latent representations, aiming to generate high-quality, semantically consistent 3D shapes under image-controlled conditions.

\begin{figure*}[!t]
\centering
\includegraphics[width=\linewidth]{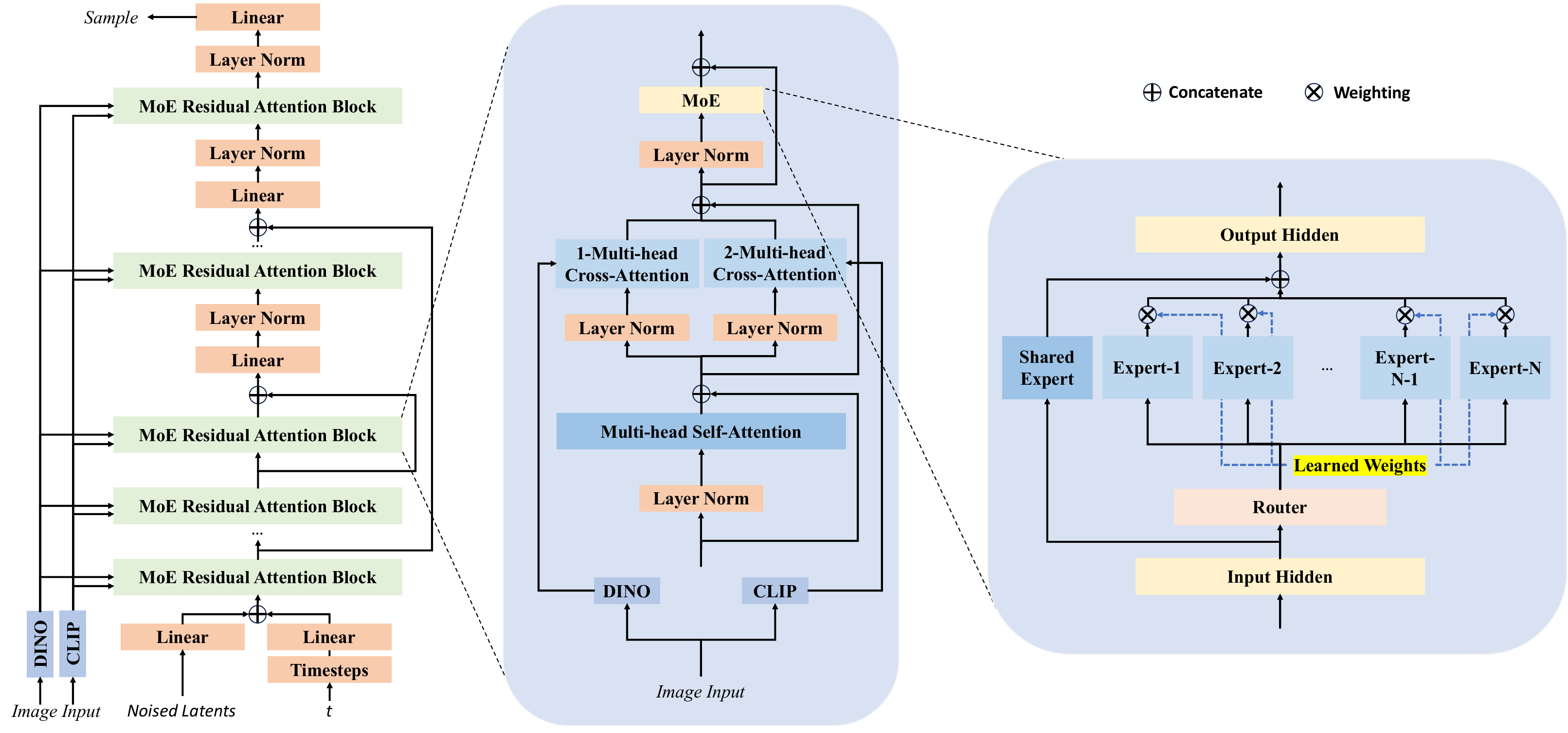}
\caption{Left: the overall architecture of \method{}. Middle: the detailed internal module of each block. Right: the detailed internal components of the MoE.}
\label{fig:TripoGen_pipeline}
\end{figure*}

\subsubsection{Image-to-3D Flow Architecture}
Our flow architecture is inspired by DiT~\cite{peebles2023scalable} and 3DShape2VecSet~\cite{zhang20233dshape2vecset}, utilizing standard transformer blocks to construct the backbone. While this architecture has demonstrated success in class-conditional tasks on ImageNet~\cite{deng2009imagenet} and ShapeNet~\cite{chang2015shapenet}, we found that naively stacking multiple transformer blocks leads to suboptimal modeling capabilities due to insufficient information fusion between shallow and deep feature. Drawing inspiration from U-ViT~\cite{bao2023all} and the UNet structure in Stable Diffusion~\cite{rombach2022high}, we follow Michelangelo~\cite{zhao2024michelangelo} by introducing long skip residual connections between blocks to capture comprehensive feature information, enhancing the network's representational capacity.

As shown on the left side of Fig.\ref{fig:TripoGen_pipeline}, the backbone is divided into three parts: encoder blocks, a middle block, and decoder blocks, with an equal number $N$ of blocks in both the encoder and decoder. Each encoder block is connected to its corresponding decoder block via skip connections. Specifically, the output of the $i$-th encoder block is skip-connected to the output of the $(N-i)$-th decoder block.
Our flow backbone is composed of $2N+1$ transformer blocks with residual connections between them. In this setup, $N$ is 10, the hidden dimension $W$ is 2048, and each transformer block has 16 attention heads.
The entire flow architecture comprises approximately 1.5 billion parameters.
The following equation describes the flow architecture with skip-connections, where $\mathtt{DB}$ denotes the decoder block and $\mathtt{EB}$ denotes the encoder block.
\begin{gather}
\begin{aligned}
\mathbf{Z}_{\mathtt{DB}}^{\left(N-i\right)}=\mathtt{{DB}}^{\left(N-i\right)}&\left(\mathbf{Z}_{\mathtt{DB}}^{\left(N-i-1\right)}\right) + \mathtt{EB}^{\left(i\right)}\left(\mathbf{Z}_{\mathtt{EB}}^{\left(i-1\right)}\right), \\
&i \in \{0,1,...,N\}.
\end{aligned}
\end{gather}
Building on \method{}'s backbone, we designed a method to inject both timestep and image conditioning, enabling controllable 3D generation. For timestep $t$ we first encode it using the $\mathtt{Timesteps}$ layer from the diffusers library\cite{von-platen-etal-2022-diffusers}, followed by an $\mathtt{MLP}$ layer that projects it to the hidden dimension $W$, obtaining a $1 \times W$ feature. 
Similarly, for the input latent $X$, with dimensions $L\times C$ encoded by the VAE, we project it to the hidden dimension $W$ using an $\mathtt{MLP}$, producing a $L\times W$ feature. Following the design of Michelangelo\cite{zhao2024michelangelo} and CLAY\cite{zhang2024clay}, we concatenate the features of timestep $t$ and latent $X$, yielding a $(L+1) \times W$ feature, which is then fed into the flow backbone.

For image conditioning, Michelangelo\cite{zhao2024michelangelo} implements conditioning by concatenating global features extracted by CLIP\cite{radford2021learning} with the input latent $X$. However, using global image features and concatenation-based injection results in a loss of fine control over the generated 3D shapes. In contrast, CLAY\cite{zhang2024clay}, which pioneered explored the large-scale 3D generation, replaces the concatenation method with a cross-attention mechanism for injecting image information. CLAY trains a 1.5 billion parameter text conditioning base model at high computational cost (256 A800 GPUs over 15 days), then freezes the base model and trains an additional 352 million parameters for image conditioning via cross-attention with reduced computational cost (over 8 hours). However, while this approach shortens the training time for image conditioning compared to text conditioning, it limits the ability to fully update model parameters based on image information, resulting in challenges in achieving fine-grained consistency between the generated 3D shapes and the image condition.
Furthermore, captions generated from rendered images can introduce semantic gaps due to lighting, shadows, and textures, which deviate from the actual 3D geometry. Even when using tools like GPT-4V\cite{openai2023gpt4v} to generate captions for 3D models, accuracy issues persist. These added noise to the alignment between captions and shapes, slowing down training convergence and posing challenges to precision.

In contrast, our approach directly leverages CLIP-ViT-L/14\cite{radford2021learning} to extract global image features $I_{\text{global}}$ and DINOv2-Large\cite{oquab2023dinov2} to extract local image features $I_\text{local}$. In each flow block, both global and local features are injected simultaneously using separate cross-attention mechanisms. The outputs are then combined with the original input and passed to the next stage. This method allows the model to attend to both global and local image information in every block, enabling faster training convergence while maintaining strong detail consistency between the generated 3D model and the input image.

The process within each block of the flow architecture can be expressed by the following equation.
\begin{gather}
\mathbf{Z}=\mathtt{Concat}\left(\mathbf{X, t}\right) \\
\mathbf{Z}=\mathbf{Z} + \mathtt{SelfAttn}\left(\mathtt{Norm}\left(\mathbf{Z}\right)\right) \\
\mathbf{Z}=\mathbf{Z} + \mathtt{CrossAttn}\left(\mathtt{Norm}\left(\mathbf{Z}\right), I_\text{local}\right) + \\ 
\mathtt{CrossAttn}\left(\mathtt{Norm}\left(\mathbf{Z}\right), I_{\text{global}}\right) \\
\mathbf{Z}=\mathbf{Z} + \mathtt{FFN}\left(\mathtt{Norm}\left(\mathbf{Z}\right)\right)
\end{gather}

\subsubsection{Rectified Flow Based Generation}
We trained the 3D generation model using our designed flow architecture, exploring sampling strategies including DDPM, EDM, and Rectified Flow, and ultimately selected Rectified Flow for the final generative model.

DDPM leverages a Markov chain to establish a connection between Gaussian noise space and the data distribution, enabling high-quality data generation. Specifically, noise $\epsilon$ is progressively added to the data $x_0$, transforming it into a standard Gaussian distribution. The data sample $x_t$ at any time step $t$ can be expressed by the following equation:
\begin{equation}\label{eq:ddpm}
x_t = \sqrt{\bar{\alpha}_t} \, x_0 + \sqrt{1 - \bar{\alpha}_t} \, \epsilon
\end{equation}
Where 
$\bar{\alpha}_t = \prod_{s=1}^t \alpha_s$, 
$\alpha_t = 1 - \beta_t$
and $\beta_t$ is the predefined noise scheduling parameter.
From the perspective of interpolation, DDPM models a relatively complex curved trajectory from $x_0$ to $x_t$.

EDM redesigns the noise schedule and sampling method, adopting a continuous-time framework to improve both the sampling speed and generation quality of DDPM. The data sample $x_t$ at any time step $t$ is modeled using the original data $x_0$ and noise $\epsilon$ as follows:
\begin{equation}\label{eq:edm}
x_t = x_0 + \sigma(t)\epsilon
\end{equation}
$\sigma(t)$ is a continuous noise standard deviation function, allowing for more flexible noise scheduling strategies, such as the power form $\left[ \left( \sigma_{\text{max}}^{1/\rho} - \sigma_{\text{min}}^{1/\rho} \right) t + \sigma_{\text{min}}^{1/\rho} \right]^\rho$, $\sigma_{\text{min}}$, where and $\sigma_{\text{max}}$ are the minimum and maximum noise standard deviation, and $\rho$ is the hyperparameter that controls the shape of the curve. 
EDM provides a more streamlined approach to modeling $x_t$ compared to DDPM. From an interpolation perspective, EDM also models a curved trajectory from $x_0$ to $x_t$.

Is there a simpler linear trajectory modeling process from $x_0$ to $x_t$? To explore this, we further investigated Rectified Flow, which learns a vector field to map the noise distribution to the data distribution. The data sample $x_t$ at any time step $t$ is modeled using the original data $x_0$ and noise $\epsilon$ as follows:
\begin{equation}\label{eq:rf}
x_t = t x_0 + (1-t)\epsilon
\end{equation}
This represents a simpler linear trajectory, offering a more efficient and streamlined approach compared to DDPM (Eq.\ref{eq:ddpm}) and EDM (Eq.\ref{eq:edm}).

Rectified flow's linear sampling simplifies network training, making it more efficient and stable, which we leverage to train our 3D flow model.
Additionally, drawing inspiration from SD3 logit-normal sampling, we increase the sampling weight for intermediate steps, as predictions for $t$ in the middle of the range $(0,1)$ are more challenging during Rectified Flow training. The sampling weight is adjusted using the following equation, where $m$ is the biasing location parameter and $s$ is the distribution width parameter.
\begin{equation}
\scalebox{0.95}{$
\pi_{\ln}(t; m, s) = \\
\frac{1}{s \sqrt{2\pi} t(1 - t)} \exp\left(-\frac{(\log(t/(1-t)) - m)^2}{2s^2}\right)
$}
\end{equation}

It is well-known that higher resolutions require more noise to sufficiently disrupt the signal. As resolution increases, the uncertainty in the noised latent at the same timestep decreases. Therefore, following SD3, we introduce \textit{Resolution-Dependent Shifting of Timestep} to adjust the timestep during both training and sampling. By remapping to a new timestep, we maintain the same level of uncertainty as with the original resolution.
We define the resolution of the first stage of our progressive training as the base resolution, denoted as 
$n$, with its timestep represented as $t_n$. The subsequent stage's resolution is defined as the fine-tune resolution, denoted as $m$, with its timestep represented as $t_m$. The relationship between $t_m$ and $t_n$ is expressed by the following equation.
\begin{equation}
t_m = \frac{\sqrt{\frac{m}{n}} t_n}{1 + \left(\sqrt{\frac{m}{n}} - 1\right) t_n}
\end{equation}

Leveraging Rectified Flow with logit-normal sampling and resolution-dependent shifting of timestep, we train our 3D flow model.

\subsection{Model and Resolution Scale-up Strategy.}\label{sec:scale_up}
Larger latent resolutions and more extensive models undoubtedly lead to performance improvements. To generate even better results, we aim to scale up both latent resolution and model size while minimizing training and inference costs. Specifically, we increased the latent resolution from 2048 to 4096, and scaled the model parameters from 1.5B to 4B using a Mixture-of-Experts (MoE).

Since the VAE training does not incorporate additional positional encoding in its input, and the varying number of query points used to learn the latent representations are downsampled from a fixed set of surface points, the VAE can generalize to resolutions beyond the training set. And higher number of query points (latent resolution) improves modeling capacity. This extrapolation ability eliminates the need for retraining the VAE, allowing us to directly encode and decode at a 4096 resolution using the VAE trained on $\{512, 2048\}$ resolutions. By leveraging this method to directly increase the latent resolution to 4096, we provide the flow model with finer geometric latent representations for training.

Additionally, to mitigate the risk of unstable training and potential loss divergence during mixed-precision training, \cite{dehghani2023scaling} recommend normalizing $Q$ and $K$ before attention operations. Following this approach, during fine-tuning at higher resolutions in our flow architecture, we apply learnable RMSNorm\cite{zhang2019root} to normalize $Q$ and $K$ within the transformer blocks.

\begin{figure}
    \centering
    \includegraphics[width=\linewidth]{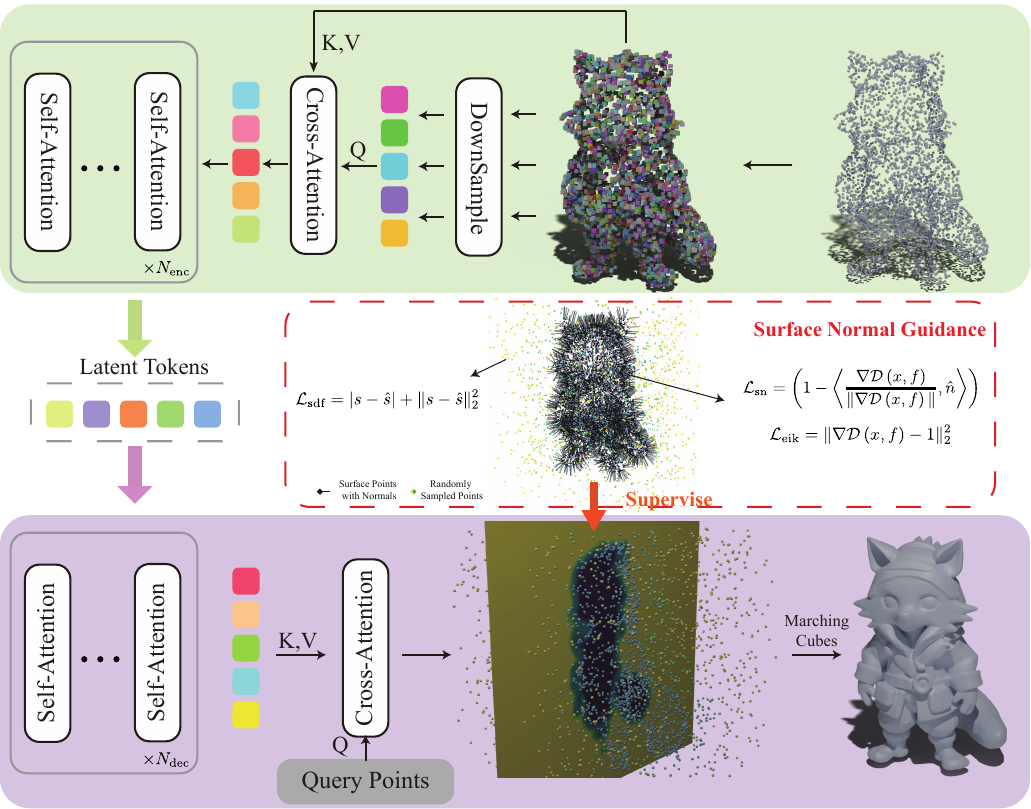}
    \caption{\method{}'s transformer-based VAE architecture. The upper is the encoder and the lower is the decoder.}
    \label{fig:vae}
\end{figure}

Directly scaling a dense model is the most straightforward approach for increasing model parameters. While this enhances performance, it significantly increases the computational resource demands and inference latency. 
Rather than this approach, we opted to scale using a Mixture-of-Experts (MoE) architecture. This method not only boosts performance by increasing model parameters but also maintains nearly constant resource usage and inference latency due to the sparse activation of the network during inference.

\begin{figure}[t]
    \centering
    \includegraphics[width=\linewidth]{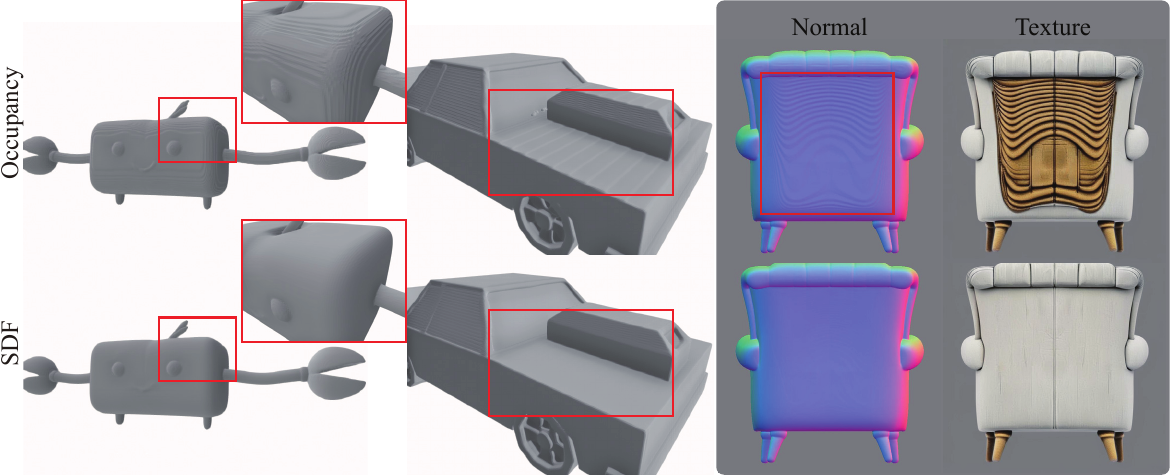}
    \caption{Comparison between model reconstruction based on Occupancy (top) and SDF (bottom).}
    \label{fig:occ-vs-sdf}
\end{figure}

As shown on the right side of Fig.\ref{fig:TripoGen_pipeline}, and following previous work\cite{riquelme2021scaling,fei2024scaling}, we extended the FFN (Feed-Forward Networks) within the transformer blocks using a Mixture-of-Experts (MoE) approach. Instead of a single $\mathtt{FFN}$ module in the dense network, $N$ parallel $\mathtt{FFN}$ expert models are employed, controlled by a gating module to scale up the model parameters. Specifically, the latent $X$ of length $L$ are distributed token-wise to different $\mathtt{FFN}$ experts by the gating module based on top-K probabilities and then reconcatenated to restore the original length $L$.
Inspired by DiT-MoE\cite{fei2024scaling}’s approach of sharing certain experts to capture common knowledge and balancing expert loss to reduce redundancy, we apply weighted activation of the top-2 experts in each token, introduce a shared expert branch across all tokens, and use an auxiliary loss to balance expert routing. 
Unlike DiT-MoE, we used the base model’s $\mathtt{FFN}$ architecture (a two-layer $\mathtt{MLP}$ with one $\mathtt{GELU}$ activation) for constructing $\mathtt{FFN}$ experts. 
Instead of training the MoE model from scratch, 
we initialized it from the base model, where the weights of the multiple $\mathtt{FFN}$ experts in each block were inherited from the corresponding $\mathtt{FFN}$ weights in the base model. Additionally, due to the shallow layers focusing on general features and the deeper layers capturing more object-specific details\cite{zeiler2014visualizing}, we limited MoE application to the final six layers of the decoder, where deep feature modeling is critical. 
This targeted scaling of parameters was applied to the most crucial part of the flow architecture. Under the action of MoE, we modified formula 17 into the following new formula.
\begin{gather}
\mathbf{Z}=\mathtt{Norm}\left(\mathbf{Z}\right), \\
\mathbf{Z}=\mathbf{Z} + \mathtt{Concat}\left(\mathtt{FFN}^{\left(i\right)}\left(\mathbf{Z}\right)\right), 
i \in \{1,2,...,N\}
\end{gather}
In our MoE scaling up, we used 8 expert models, activating the top 2 $\mathtt{FFN}$ experts per MoE block while sharing one $\mathtt{FFN}$. Additionally, MoE expansion was applied to the final 6 layers of the decoder, increasing the overall model parameters from 1.5B to about 4B.

\subsection{3D Variational Autoencoder (VAE)}\label{sec:vae}
\subsubsection{3D Model Representation}
Most existing 3D shape generation works~\cite{zhang2024clay,li2024craftsman} adopt occupancy field or semi-continuous occupancy~\cite{wu2024direct3d} as the neural implicit representation for 3D model. For each query position $x \in \mathbb{R}^3$, these methods utilize a neural network $\mathcal{D}$ to predict the occupancy value $o$ from the latent features $f$, supervised by the ground-truth occupancy value $\hat{o}$ with Binary Cross Entropy (BEC) loss: 
\begin{gather}
    o=\mathcal{D}\left(x, f\right), \\
    \mathcal{L} = \mathbb{E}_{x\in \mathcal{R}^3}\left[\text{BCE}\left(o, \hat{o}\right)\right].
\end{gather}
Learning geometry through occupancy representation as a classification task is easier to train and converge compared to the signed distance function (SDF) as a regression task.
However, the occupancy representation has limited geometric representation capabilities compared to SDF, which provides more precise and detailed geometry encoding.
Additionally, models reconstructed using occupancy representation often exhibit noticeable aliasing artifacts and typically require further post-processing (e.g., smooth filter or super-sampling) to address these issues.
Without post-processing, these aliasing artifacts sometimes also impact the subsequent texture generation.
Fig.\ref{fig:occ-vs-sdf} shows some examples of geometry reconstruction and texture generation results based on occupancy and SDF, respectively.

Given these considerations, we adopt neural SDF as our 3D model representation. This method, built upon a set of latent tokens, provides a stronger geometric detail than occupancy-based approaches~\cite{zhang2024clay,li2024craftsman,wu2024direct3d}.
Specifically, we predict the SDF value $s$ of each query position as:
\begin{equation}
    s = \mathcal{D}\left(x, f\right).
\end{equation}
For efficiency, we employ the truncated signed distance function (TSDF) in our VAE model. In the following paragraph, we use $s$ to represent TSDF for simplicity.

\subsubsection{Geometry Learning with Surface Normal Guidance}
More importantly, SDF representation theoretically ensures the effectiveness of supervision in the gradient domain of the neural implicit field. 
We think geometric details are relevant to the gradient domain of the neural implicit field, which represents the higher-order information compared to the value domain of the implicit field.
Therefore, we apply surface normal guidance during VAE training to capture finer-grained geometric details, providing a better latent space for model sampling.
Specifically, in addition to the commonly used SDF loss, our approach also includes direct supervision of finer detailed geometry learning using the ground-truth surface normals and an additional eikonal regularization:
\begin{gather}
\mathcal{L}_{\text{vae}}=\mathcal{L}_{\text{sdf}}+\lambda_{\text{sn}}\mathcal{L}_{\text{sn}}+\lambda_{\text{eik}}\mathcal{L}_{\text{eik}}+\lambda_{\text{kl}}\mathcal{L}_{\text{kl}}, \\
\mathcal{L}_{\text{sdf}}=|s - \hat{s}| + \|s - \hat{s}\|_2^2 , \\
\mathcal{L}_{\text{sn}}=\left(1-\left<\frac{\nabla \mathcal{D}\left(x, f\right)}{\|\nabla \mathcal{D}\left(x, f\right)\|}, \hat{n}\right>\right), \\
\mathcal{L}_{\text{eik}}=\|\nabla \mathcal{D}\left(x, f\right)-1\|^2_2,
\end{gather}
where $\hat{n}$ is the ground-truth surface normal, $\left<\cdot,\cdot\right>$ denotes the cosine similarity of two vectors, $\mathcal{L}_\text{eik}$ is the eikonal regularization, and $\mathcal{L}_\text{kl}$ represents the KL-regularization in the latent space.
Unlike SDF loss, which involves sampling points near the surface and randomly throughout space, the surface-normal loss is applied exclusively to surface points, making it a more efficient method for supervising fine-grained geometry learning.

\subsubsection{Network Architecture}
Following the design of 3DShape2Vecset\cite{zhang20233dshape2vecset}, we choose the latent vector set as our latent representation, which encodes a point cloud into latent space and subsequently decodes a geometry function (i.e., SDF) from it.
To facilitate more efficient scaling up, we adopt a state-of-the-art transformer-based encoder-decoder architecture~\cite{zhang20233dshape2vecset,zhang2024clay,zhao2024michelangelo}.
Specifically, we choose the downsampled version in
3DShape2Vecset~\cite{zhang20233dshape2vecset} that subsamples $M$ points $\mathbf{X}'$ from the full set of surface points $\mathbf{X}$, and directly utilizes the point cloud itself as initial latent queries instead of learnable embeddings.
Then, the surface points information, encoded by concatenating positional embedding and surface normal, is integrated into latent queries via cross-attention, resulting in compact latent tokens $\mathbf{Z}$ rich in geometric information, as shown in the following:
\begin{gather}
\mathbf{Z_0}=\mathtt{CrossAttn}\left(\mathtt{PosEmb}\left(\mathbf{X}\right),\mathtt{PosEmb}\left(\mathbf{X}'\right)\right), \\
\mathbf{Z} = \mathtt{Linear}\left(\mathtt{SelfAttn}^{\left(i\right)}\left(\mathbf{Z_0}\right)\right), i \in \{0,1,...,L_{\text{enc}}\},
\end{gather}
where $\mathtt{CrossAttn}$ denotes a cross-attention layer, $\mathtt{SelfAttn}^{\left(i\right)}$ denotes the self-attention layers, and \text{Linear} is a linear layer.

After obtaining the latent representation, we can decode the signed distance value for each query position $x \in \mathbb{R}^3$:
\begin{gather}
\mathbf{\widetilde{Z}}=\mathtt{SelfAttn}^{\left(i\right)}\left(\mathtt{Linear}\left(\mathbf{Z}\right)\right), i \in \{0, 1, ..., L_{\text{dec}}\}, \\
s=\mathtt{CrossAttn}\left(\mathtt{PosEmb}\left(x\right), \mathbf{\widetilde{Z}}\right).
\end{gather}
Finally, the mesh of the 3D model can be extracted by applying Marching Cubes~\cite{DBLP:conf/siggraph/LorensenC87} at a given resolution.

To implement progressive flow model training for faster convergence, we follow \cite{zhang2024clay} to adopt a multi-resolution VAE with $M \in \{512, 2048\}$ tokens, where the VAE weights are shared across different resolutions.
This training strategy, combined with the position-encoding-free feature of the VAE transformer, provides the VAE with strong extrapolation capabilities, allowing it to directly infer the 3D models with higher-resolution (e.g., 4096) tokens without requiring additional fine-tuning.
Unlike previous works, which used only a few surface points (only 2048 or 8192 points)\cite{zhang2024clay} as the VAE input, we opted to use a denser surface point for each 3D model.
We think the purpose of the VAE is to capture as much geometric information of the 3D model as possible, rather than functioning as a sparse point cloud reconstruction task.
The more input points provided, the more geometric information is encoded in the latent space, resulting in higher-quality geometry being decoded.

\begin{figure}[!t]
\centering
\includegraphics[width=\linewidth]{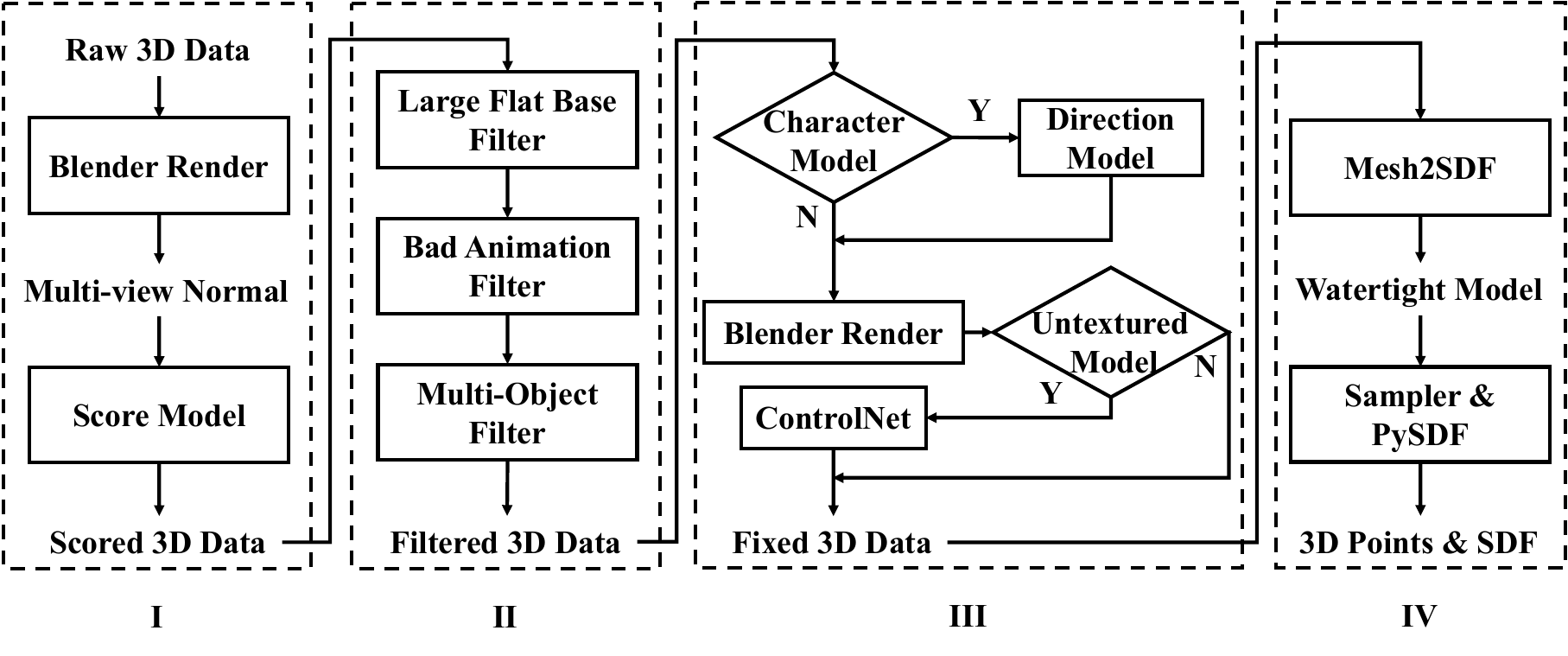}
\caption{Demonstration of the \method{} data-building system. I: Data scoring procedure; II: Data filtering procedure; III: Data fixing and augmentation procedure. IV: Field data producing procedure.}
\vspace{-1em}
\label{fig:data_process_pipeline}
\vspace{-1em}
\end{figure}

\section{Data-Building System.}\label{sec:data}
\method{} is trained on existing open-source datasets such as Objaverse (-XL)\cite{deitke2023objaverse, deitke2024objaverse} and ShapeNet\cite{chang2015shapenet}, which contains approximately $10$ million 3D data. Since most of these data are sourced from the Internet, their quality varies significantly, requiring extensive preprocessing to ensure suitability for training.
To overcome these challenges, \method{} developed a dedicated 3D data processing system that produces high-quality, large-scale datasets for model training.
As illustrated in Fig.\ref{fig:data_process_pipeline}, the system comprises four processing stages (Data Process I$\sim$IV), responsible for data scoring, filtering, fixing and augmentation, and field data producing, respectively.

\subsection{I: Data Scoring}
Each 3D model is scored, with only the high-score models advancing to the subsequent processing stages.
Specifically, we randomly selected approximately $10K$ 3D models and used Blender to render four different views of normal maps for each model. These multi-view normal maps are then manually evaluated by 10 professional 3D modelers, assigning scores on a scale from $1$ (lowest) to $5$ (highest).
Using this annotated data, we trained a linear regression-based scoring model concatenating their CLIP\cite{radford2021learning} and DINOv2\cite{oquab2023dinov2} features as input. This model was subsequently used to infer quality scores from the multi-view normal maps of all 3D models for filtering.

\subsection{II: Data Filtering}
After scoring, further filtering is applied to exclude models with large planar bases, rendering errors in animations, and those containing multiple objects.
Specifically, models with large planar bases are filtered by determining if different surface patches can be classified as a single plane, based on features composed of their centroid positions, normal vectors, and the area of the resulting plane. Blender identifies animated models, sets them to the first frame, and filters out any models that still exhibit rendering errors after being set. And models containing multiple objects are filtered by evaluating the proportion of the largest connected component on the opaque mask, along with the magnitude of the solidity of both the largest connected component and the entire mask.

\subsection{III: Data Fixing and Augmentation}
After data filtering, we perform the orientation fixing of character models to ensure they face forward. Specifically, we select 24 orientations around the x, y, and z axes, and for each, render images from six orthogonal views: front, back, left, right, top, and bottom. The DINOv2\cite{oquab2023dinov2} features from these six views are concatenated to train an orientation estimation model, which is then used to infer and fix the orientation of all character models. Additionally, for all untextured models, we render multi-view normal maps and use ControlNet++\cite{li2024controlnet++} to generate corresponding multi-view RGB data, which serve as conditional inputs during training.

\subsection{IV: Field Data Production}
Although Objaverse (-XL)~\cite{deitke2023objaverse,deitke2024objaverse} contains a large amount of data, most of the models are unsuitable for direct training, even after processing steps such as scoring, filtering, and fixing. 
Since we adopt the neural implicit field as our 3D model representation, it's necessary to convert the original non-watertight mesh to watertight ones for computing geometry supervision (e.g., occupancy or SDF). 
Rather than using common methods like TSDF-fusion~\cite{DBLP:conf/ismar/NewcombeIHMKDKSHF11} or ManifoldPlus~\cite{DBLP:journals/corr/abs-1802-01698,DBLP:journals/corr/abs-2005-11621}, we are inspired by \cite{DBLP:journals/tog/WangLT22,zhang2024clay} to construct a Unsigned Distance Function (UDF) field with a resolution of $512^3$ grid from the original non-watertight mesh, and then apply Marching Cubes~\cite{DBLP:conf/siggraph/LorensenC87} to extract the iso-surface with a small threshold $\tau=\frac{3}{512}$. 
To remove interior structure for more efficient geometry learning, we follow \cite{zhang2024clay} by resetting the UDF value of the invisible grids to prevent the extraction of interior iso-surface before applying Marching Cubes. 
We then remove some small and invisible interior mesh components by calculating the area and the ambient occlusion ratio of each mesh component.
Finally, we uniformly sample surface points along with their normals, and randomly sample points both within the volume and near the surface.

\section{Experiments}

\subsection{Implementation Details}\label{sec:Implementation_detail}

\begin{figure*}[!t]
    \centering
    \includegraphics[width=1\linewidth]{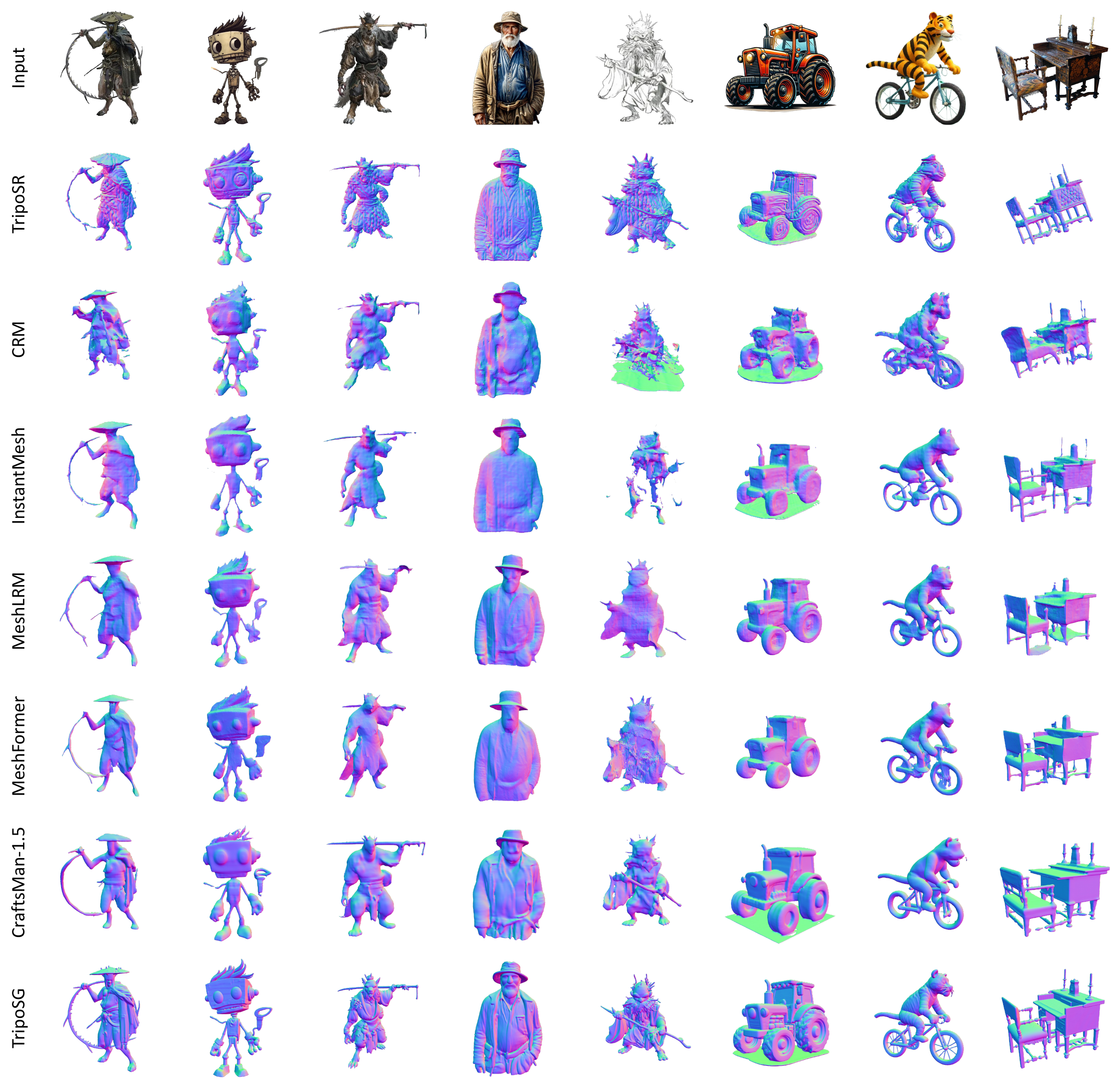}
    \vspace{-2.5em}
    \caption{Comparison of 3D generation performance of \method{} and other previous state-of-the-art methods under the same image input.}
    \label{fig:demo_comparison}
    \vspace{-1em}
\end{figure*}

The shape generation experiments are divided into two parts.
In the \method{} experiment, we progressively scaled both resolution and model size.
First, we trained a 1.5B parameter model on a 2M dataset with a latent resolution of 512 tokens, using a learning rate of 1e-4 for 700k steps. 
Next, we switched to a latent resolution of 2048 tokens and continued training for an additional 300k steps with a learning rate of 5e-5. 
Finally, to scale up, we expanded the model parameters to 4B using MoE and increased the latent resolution to 4096 tokens. 
Training resumed on 1M high-quality dataset with a learning rate of 1e-5 for 100k steps. 
The batch size of the three processes is set to 16, 10, and 8 per GPU respectively.
The entire training process took approximately 3 weeks across 160 A100 GPUs.

In the ablation experiments, 
we still use a small dataset (180K) filtered from Objaverse and a 975M parameter model for training. For the non-scaling ablation experiments, as shown in Tab.\ref{tab: diffusion_improvements_abaltion}, we trained the model with a latent resolution of 512 tokens, a learning rate of 1e-4, for about 300k steps, over approximately 3 days on 32 A100 GPUs. For the scaling-up ablation experiments, as shown in rows 2-4 of Tab.\ref{tab: diffusion_scaling_abaltion}, we progressively continued training from the previous experiment with latent resolutions of 2048 tokens, 4096 tokens, and 4096 tokens with the MoE model architecture, respectively, for an additional 100k steps. The learning rates were 5e-5, 1e-5, and 1e-5, respectively. These three scaling-up experiments took around 9 days in total on 32 A100 GPUs. For all ablation experiments, the batch size was set to 16 per-GPU.

It is also worth mentioning that during training, the image foreground is resized to a fixed ratio (90\%) and rotated around the center within a range of $[-10^\circ, 10^\circ]$ with a probability of 0.2. This setting enables the model to generalize well to various input images. During inference, the image is first detected for the foreground and then resized to the same ratio as the training foreground to obtain the best generation effect.

\begin{figure}[t]
    \centering
    \includegraphics[width=1\linewidth]{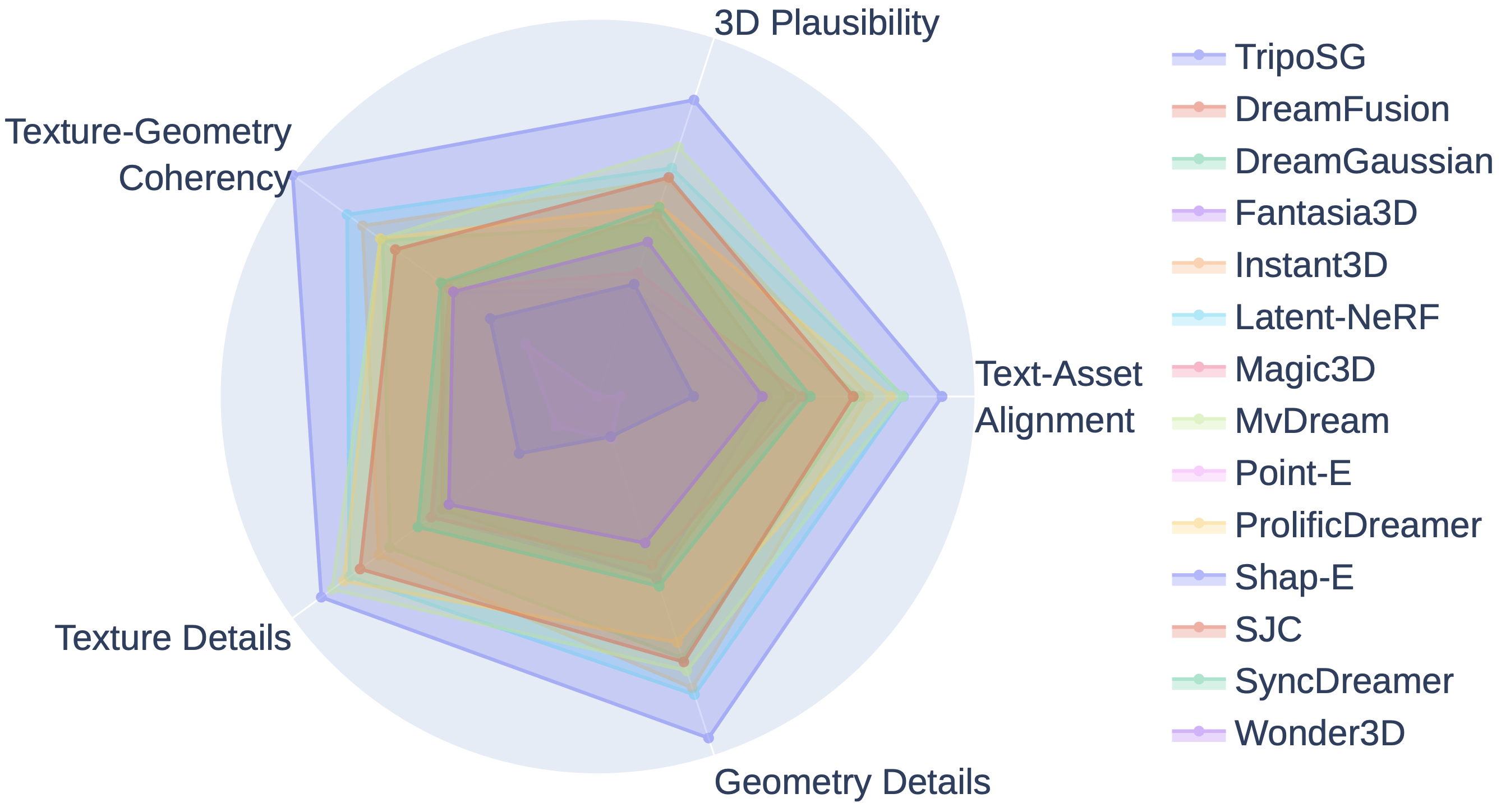}
    \caption{Radar chart of the score of different methods in 5 aspects, including 3D plausibility, text-asset alignment, geometry details, texture details, texture-geometry coherency.}
    \label{fig:gpteval}
    \vspace{-1em}
\end{figure}

Following ~\cite{zhao2024michelangelo}, our VAE model adopts a network architecture with an 8-layer encoder and a 16-layer decoder. We use a larger decoder to enhance the ability to decode geometry from the latent space, without increasing the inference cost of the VAE during the flow model training stage.
The weights of surface normal loss $\lambda_{\text{sn}}$, eikonal regularization $\lambda_{\text{eik}}$, and KL-regularization $\lambda_{\text{kl}}$ are set to $10$, $0.1$, and $0.001$, respectively.
For each training data item, our model takes $20,480$ surface points as input and randomly samples $8,192$ near-surface points, $8,192$ volume points, and $8,192$ on-surface points for supervision.

The VAE experiments are divided into two parts as well: \method{} experiments and ablation experiments.
In the \method{} experiment, we train the VAE via SDF supervision with surface normal guidance and eikonal regularization using a learning rate of 5e-5 and a per-GPU batch size of 6 for 2.5M steps. The training process takes approximately 12 days on 32 A100 GPUs, then the VAE is used for the scale-up flow model training.
For the ablation experiment, we evaluate the VAE reconstruction quality from different experiment settings on a small dataset (180K filtered data from Objaverse). We train the VAE with different settings using a learning rate of 1e-4 and a per-GPU batch size of 8 for 286K steps on 8 A100 GPUs.

\subsection{Dataset, Metrics and Baselines}
\subsubsection{Dataset}
We train our model on Objaverse (-XL)~\cite{deitke2023objaverse,deitke2024objaverse}, the largest publicly available 3D dataset, which contains over 10 million unique 3D objects from various sources.
Since most of the data in Objaverse(-XL) cannot be directly used for our model's training, we apply the preprocessing steps introduced in Sec.\ref{sec:data} to prepare the training data, including scoring, filtering, orientation fixing, and training data reproducing.
After preprocessing, we get 2 million high-quality 3D objects. We compute the ground-truth SDF from sampled points based on the reproduced 3D models.
For single-image conditioned flow model training, we render from 8 random viewpoints in front of the 3D models with randomly sampled parameters from the camera's focal length, elevation, and azimuth ranges after orientation fixing but before data reproducing. Specifically, the range of elevation is $[-15^\circ, 30^\circ]$. 
The range of azimuth is $[0^\circ, 180^\circ]$.
The range of focal length is randomly selected from a discrete focal length list. The focal length list is [orthogonal, 50mm, 85mm, 135mm, 2 randomly selected from 35mm-65mm].

\subsubsection{Metrics}\label{sec:metric}
For flow model generation quality, we use FID\cite{heusel2017gans} to evaluate our model performance.
The original goal of FID is to evaluate the quality and photorealism of the generated shapes. We aim to introduce FID-related metrics to quantitatively assess the quality of the 3D models generated by \method{}. Typically, the input for generating 3D models is a 2D RGB image, while \method{} primarily generates textureless 3D models, creating an evaluation gap between the two.
To bridge this gap and enable evaluation on a consistent semantic level, we propose an improved evaluation process. Specifically, for 3D ground-truth models, we render paired RGB images $I_{gt}$ and normal maps $N_{gt}$ under the same viewpoints. The RGB images $I_{gt}$ are input into \method{} to generate 3D shapes, from which we render normal maps $N_{gen}$ from the same viewpoints as the input images. We then compute the Normal-FID between the generated normal maps $N_{gen}$ and the ground-truth normal maps $N_{gt}$ to evaluate the overall performance of \method{}.
In addition, the existing metrics, even the Normal-FID, are not flexible enough to adapt to various evaluation criteria and may not accurately reflect human preferences.
Thus, we also incorporate a recently introduced evaluation metric GPTEval3D~\cite{DBLP:conf/cvpr/WuYLZLGLW24} for further comparison.

We adopt some commonly used metrics to evaluate VAE reconstruction quality and flow model generation quality.
For VAE reconstruction quality, we focus on the accuracy of the reconstruction mesh from input points (with corresponding normals). We use the Chamfer distance, F-score with $0.02$ threshold, and the normal consistency as metrics.

\subsection{Quantitative and Qualitative Evaluation\protect\footnote{The original images in Fig.\ref{fig:teaser}, Fig.\ref{fig:demo_comparison}, Fig.\ref{fig:ablation_vis} and Fig.\ref{fig:geo_demo_show}, Fig.\ref{fig:texture_demo_show} are sourced from various 3D generation platforms, benchmarks (such as Rodin, Meshy, 3D Arena), and our own collections.}}

\subsubsection{Comparison with Different Methods in Visualization}
As shown in Fig.\ref{fig:demo_comparison}, we compare \method{} with the most popular image-to-3D generation methods\cite{tochilkin2024triposr, wei2024meshlrm, xu2024instantmesh,wang2024crm, li2024craftsman, DBLP:journals/corr/abs-2408-10198}. It is worth noting that for Craftsman\cite{li2024craftsman}, we used the online demo of Craftsman-1.5 on Huggingface for inference, which is a more advanced version of the Craftsman. 
The first row in the figure shows the original input image, while rows 2-7 present a comparison between the generation 3D models of other methods and \method{}. We compared their geometric quality by rendering normal maps. The results shown in the figure are all 3D normal maps rendered from the same viewpoint.
Notably, we preprocessed the original image by removing the background and fed the processed images to different open-source models via Huggingface demos for online inference and generation. Unlike previous works that typically compare 3D generation results on simple, standard images, we conducted comparisons on complex and widely varying cases.

Specifically, we evaluated the methods across five dimensions from left to right: (1) Semantic Consistency: \method{} generates 3D models with better semantic consistency, as shown in the first and second cases, with greater detail and semantic alignment. (2) Detail: The third and fourth cases demonstrate \method{}'s ability to capture finer details, such as clothing textures and accessories, providing richer visual fidelity. (3) Generalization: The fifth and sixth cases highlight \method{}'s ability to generate high-quality 3D models from both comic-style and cartoon-style images, showcasing its strong generalization capability. (4) Spatial Structure Generation: The seventh and eighth cases show \method{} excels at generating complex spatial structures, demonstrating superior spatial modeling capabilities. (5) Overall Performance: We compared \method{} with the latest and most advanced open-source methods, including both reconstruction and generation approaches, and it is evident that \method{} delivers significantly superior results, leaving a strong impression and outperforming previous approaches by a wide margin.

\subsubsection{Comparison with Different Methods in Metric}

Benefiting from the development of Large Multimodal Models (LMMs), we can easily leverage them to obtain evaluation results more aligned with human preferences. We adapt the evaluation script and test text prompt from \cite{DBLP:conf/cvpr/WuYLZLGLW24} and use the Claude3.5 instead of GPT-4Vision~\cite{DBLP:journals/corr/abs-2303-08774} as the Large Multimodal Models (LMMs).
We compare our results with various types of previous SOTA methods\cite{  
DBLP:conf/iclr/PooleJBM23, 
DBLP:conf/iclr/TangRZ0Z24, 
DBLP:conf/iccv/ChenCJJ23, 
li2023instant3d, 
DBLP:conf/cvpr/MetzerRPGC23, 
DBLP:conf/cvpr/Lin0TTZHKF0L23, 
DBLP:conf/iclr/ShiWYMLY24, 
DBLP:journals/corr/abs-2212-08751, 
DBLP:conf/nips/Wang00BL0023, 
DBLP:journals/corr/abs-2305-02463, 
DBLP:conf/cvpr/WangDLYS23, 
DBLP:conf/iclr/LiuLZLLKW24, 
long2024wonder3d}.
Specifically, we use an off-the-shelf text-to-image model Flux\cite{flux} to generate the input image for our method. 
Fig.\ref{fig:gpteval} demonstrates a Radar chart comparing the evaluation results across five aspects for different methods, as assessed by the LMM model.
The result indicates that our \method{} outperforms the other methods in all aspects.

\subsubsection{SOTA Performance of \method{}}
\begin{figure*}
    \centering
    \includegraphics[width=1\linewidth]{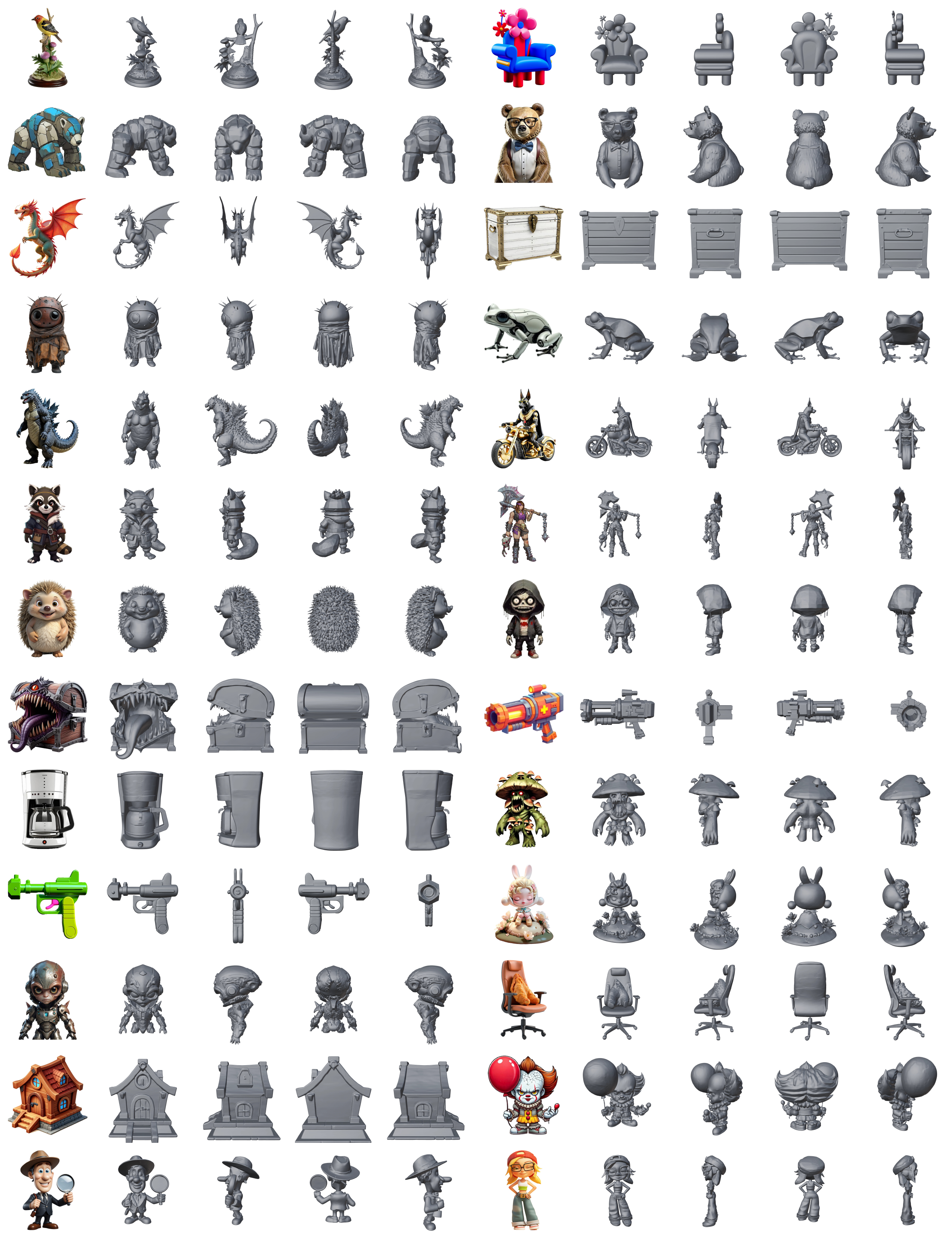}
    \caption{A diverse array of texture-free 3D shapes generated by \method{}.}
    \label{fig:geo_demo_show}
\end{figure*}

\begin{figure*}
    \centering
    \includegraphics[width=1\linewidth]{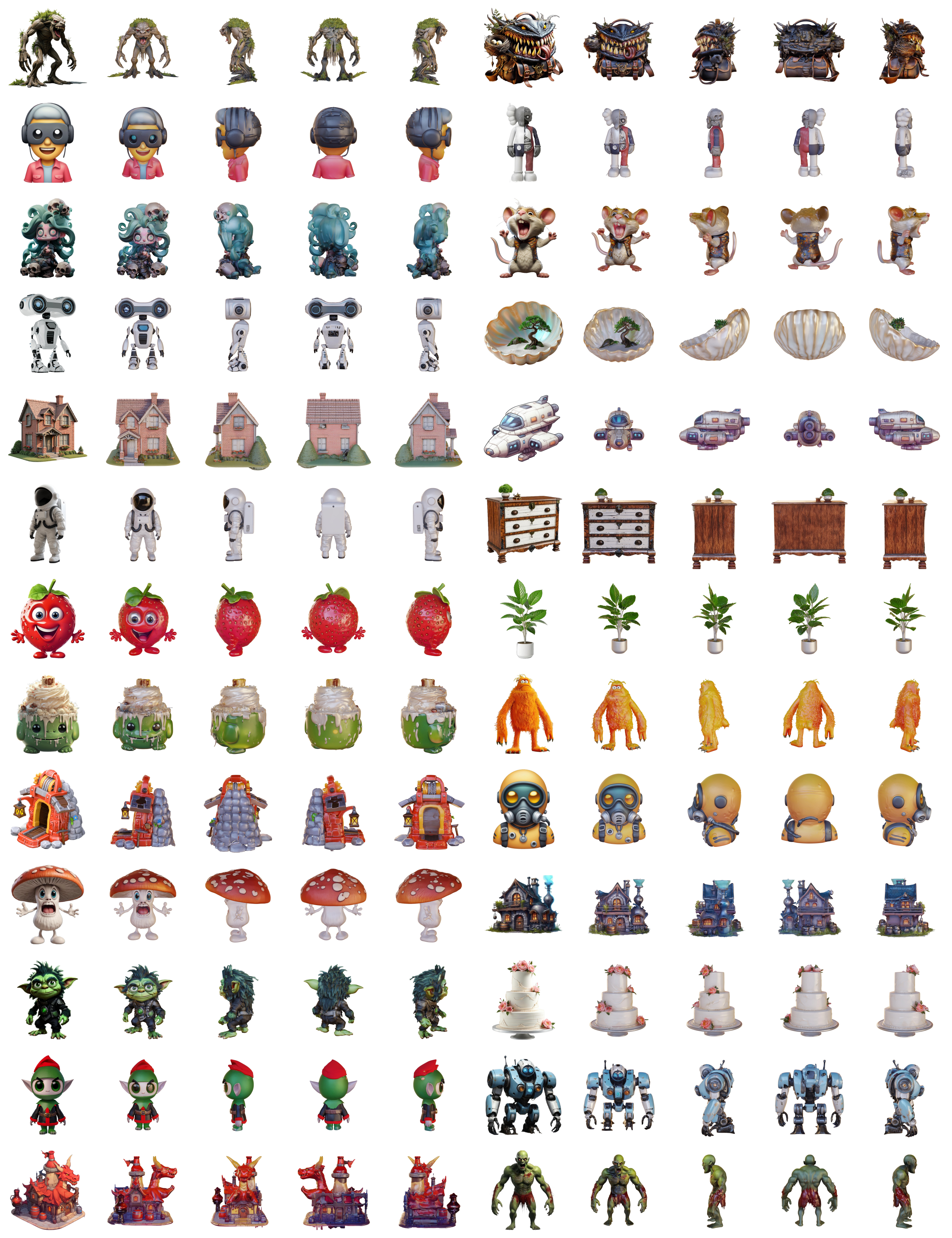}
    \caption{A diverse array of textured 3D shapes generated by \method{}.}
    \label{fig:texture_demo_show}
\end{figure*}

Fig.\ref{fig:teaser}, Fig.\ref{fig:geo_demo_show} and Fig.\ref{fig:texture_demo_show} showcase various image-to-3D results generated by \method{}. Notably, there are no duplicates among these cases, and the generated models have not undergone any post-processing (such as smoothing or removing floaters). Textured cases were produced through texture map generation, while non-textured cases were rendered from the original mesh. The process of texture generation is detailed in Sec.\ref{sec:texture}. The first image of each case in Fig.\ref{fig:geo_demo_show} and Fig.\ref{fig:texture_demo_show} is the input image, and the following four images are multi-view results rendered from the generated 3D model.

From these results, it is evident that \method{} delivers outstanding 3D model generation. Across the wide range of showcased cases—covering various complex structures, diverse styles, imaginative designs, multi-object compositions, thin surfaces, and richly detailed scenarios—\method{} consistently produces impressive 3D models. Achieving this level of performance is challenging for existing methods. The strong generalization highlights the advantages of large-scale datasets, while the rich detail and interpretive capability underscore the benefits of our high latent resolution and large model size, collectively reflecting \method{}’s state-of-the-art performance.

\section{Texture Generation}\label{sec:texture}
Thanks to the finely detailed and high-quality 3D geometry generated by \method{}, referring to Meta 3D TextureGen~\cite{bensadoun2024meta}, we can leverage the rendered normal maps as input conditions for existing mature multi-view generation methods to produce consistent multi-view texture images. These multi-view texture images are then projected onto the geometric surface to obtain detailed texture maps. 
Fig.\ref{fig:texture_demo_show} shows the 3D result with texture maps generated by \method{}.
\section{Ablation and Analysis}

\begin{table}[t]
\begin{center}
\caption{The ablation for flow model improvements. `Skip-C' is the skip-connection operation and `Sample-S' is the sample schedule.}
\vspace{1em}
\label{tab: diffusion_improvements_abaltion}
\begin{sc}
\resizebox{0.48\textwidth}{!}{
\begin{tabular}{| c | c | c| c|}
\hline
Condition & Skip-C & Sample-S & Normal-FID $\downarrow$\\
\hline
Dinov2 & \ding{55} & R-Flow  & 10.69\\
\hline
\multirow{4}{*}{CLIP-Dinov2} & \ding{55} & R-Flow  & 10.61\\
\cline{2-4}
 & \ding{51} & DDPM  & 9.63\\
\cline{2-4}
 & \ding{51} & EDM  & 9.50\\
\cline{2-4}
 & \ding{51} & R-Flow  & 9.47 \\
\hline
\end{tabular}
}
\end{sc}
\end{center}
\vspace{-2em}
\end{table}
\subsection{Ablation for Flow Model}

To validate the effectiveness of the proposed flow model improvements and scaling-up strategies, we performed specific ablation experiments and comparative analyses for each improvement.
Using a further filtered $180K$ high-quality Objaverse dataset, we conducted ablation experiments following the training settings in \ref{sec:Implementation_detail}, and evaluated the results using the Normal-FID metric introduced in \ref{sec:metric}.

For the evaluation of Normal-FID, we selected $1K$ data samples from the $180K$ dataset as a dedicated test set for 3D generation performance validation, with the remaining data samples used for training. For the test set, we rendered each 3D ground-truth model's front view paired RGB and normal map using a $50mm$ camera focal length and a $10^\circ$ elevation (the test set rendering settings are included within the training set settings). The RGB images are used to generate 3D shapes, and the normal maps are compared with the normal maps rendered from the generated 3D shapes to calculate the Normal-FID.

Our flow model ablation experiment consists of two parts: flow model improvement training and flow model scaling up. For the flow model improvement experiments, as shown in Tab.\ref{tab: diffusion_improvements_abaltion}, we used a $975M$ parameter model with a latent resolution of $512$ tokens and trained for $300K$ steps for each experiment on the high-quality Objaverse dataset. We conducted comparative analyses on $Condition$, $Skip-connection$, and $Sampling-schedule$ improvements.
From the last three rows in Tab.\ref{tab: diffusion_improvements_abaltion}, we observe that R-Flow sampling yields better generation results compared to EDM and DDPM. Combined with its training efficiency, R-Flow demonstrates clear advantages in 3D generation tasks. Comparing rows 2 and 5 shows that the skip-connection operation significantly affects generation results, with the fusion of deep and shallow features improving flow modeling. Additionally, the comparison between the first two rows indicates that the CLIP condition also slightly improves generation results. 
From the overall quantitative results, the skip-connection operation has the most obvious effect among these ablations.

\begin{table}[t]
\begin{center}
\caption{The ablation for flow model scaling up.}
\vspace{1em}
\label{tab: diffusion_scaling_abaltion}
\begin{sc}
\resizebox{0.48\textwidth}{!}{
\begin{tabular}{| c| c|c|c|}
\hline
Dataset  & Token number & MoE &Normal-FID $\downarrow$\\
\hline
\multirow{4}{*}{Objaverse}  &512 & \ding{55}& 9.47 \\
\cline{2-4}
  &2048 & \ding{55}& 8.38\\
\cline{2-4}
  &4096 & \ding{55}& 8.12\\
\cline{2-4}
  &4096 & \ding{51} & 7.94\\
\hline
\method{}  &4096 & \ding{51} & 3.36\\
\hline
\end{tabular}
}
\end{sc}
\end{center}
\vspace{-2em}
\end{table}

For the flow model scaling-up experiments, as shown in rows 2-4 of Tab.\ref{tab: diffusion_scaling_abaltion}, we used a $975M$ parameter model with CLIP-DINOv2 dual-conditioning, skip-connection operations, and rectified-flow sampling schedule. These models are trained for a total of $300K$ steps on the high-quality Objaverse data to conduct comparative analyses on latent resolution and MoE. The last row of Tab.\ref{tab: diffusion_scaling_abaltion} represents our largest \method{} model, encompassing the largest data, model size, resolution, and training cost.
From the first three rows of Tab.\ref{tab: diffusion_scaling_abaltion} we observe that as the latent resolution increases, the generated results consistently improve, with the most significant improvement occurring from 512 to 2048 tokens. Comparing rows 3 and 4 shows the gains from increasing model parameters through MoE. 
Comparing rows 4 and 5 demonstrates the performance improvement achieved by increasing high-quality data size. When combined with the results from row 1, we can see that the improvement from the increased high-quality data size surpasses that from higher resolution.
Overall, the large-scale dataset, large model size, and high resolution contribute to significant performance improvements, allowing \method{} to achieve remarkable 3D generation results.


\subsection{Ablation for VAE}

\begin{table}[t]
\begin{center}
\caption{The ablation of different VAE, including 3D representation, training supervision and training dataset. `Repr' refers to the type of 3D representation used, and `$\mathcal{L}_\text{sn}$' and `$\mathcal{L}_\text{eik}$' refer to surface normal loss and eikonal regularization respectively. The `Dataset' indicates whether a large (TripoSG) or a small dataset (Objaverse) is used. }
\vspace{1em}
\label{tab:vae}
\begin{sc}
\resizebox{0.48\textwidth}{!}{
\begin{tabular}{| c | c c c | c c c |}
\hline
Dataset & Repr. & $\mathcal{L}_\text{sn}$ & $\mathcal{L}_{eik}$ & Chamfer $\downarrow$ & F-score $\uparrow$ & N.C. $\uparrow$ \\
\hline
\multirow{4}{*}{Objaverse} & Occ & \ding{55} & \ding{55} & 4.59 & 0.999 & 0.952 \\
\cline{2-7}
 & SDF & \ding{55} & \ding{55} & 4.60 & 0.999 & 0.955 \\ 
\cline{2-7}
 & SDF & \ding{51} & \ding{55} & 4.56 & 0.999 & 0.956 \\
\cline{2-7} 
 & SDF & \ding{51} & \ding{51} & 4.57 & 0.999 & 0.957 \\
\hline 
\method{}& SDF & \ding{51} & \ding{51} & 4.51 & 0.999 & 0.958 \\
\hline 
\end{tabular}
}
\end{sc}
\end{center}
\end{table}

\begin{figure}[t]
    \centering
    \includegraphics[width=1\linewidth]{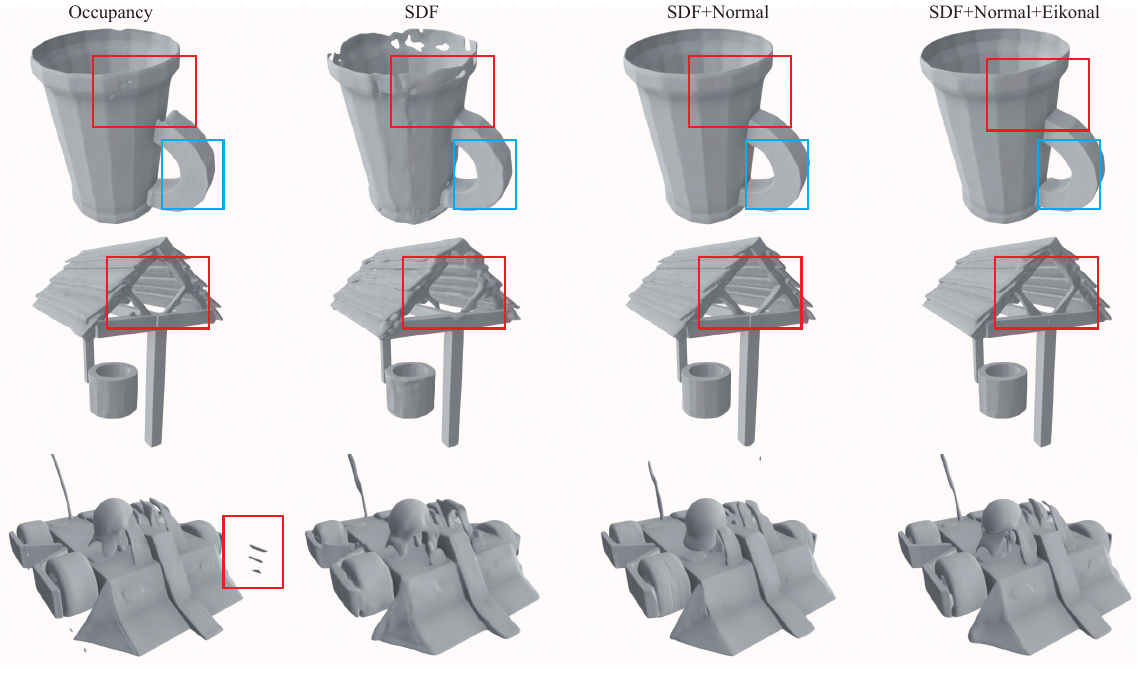}
    \caption{Qualitative comparison for the ablation of VAE with different types of 3D representation and training supervision.}
    \label{fig:vae-comparison}
    \vspace{-1em}
\end{figure}

To evaluate the effectiveness of neural SDF implicit representation with surface normal guidance, we experiment with different VAE model settings, including the formulation of neural implicit representation, training supervision, and training dataset.
Tab.\ref{tab:vae} demonstrates the qualitative results of VAE reconstruction quality with different training settings.
We can observe that the SDF representation, combined with surface normal guidance and eikonal regularization, improves the reconstruction quality and geometry details, achieving lower Chamfer distance and higher normal consistency compared to occupancy-based results.
As the amount of training data increases (as demonstrated by \method{}), the reconstruction quality of the VAE further improves.
Fig.\ref{fig:vae-comparison} provides qualitative comparisons between them.
Occupancy-based reconstruction results suffer from aliasing artifacts (highlighted by the blue box), thin structures, and floaters (highlighted by the red boxes).
While SDF representation avoids aliasing artifacts, there remains a gap in achieving high-quality reconstruction, particularly for thin-shell structures where performance may worsen. 
Incorporating surface normal guidance can result in sharper reconstructions with finer details. However, over-emphasizing surface normal guidance during training introduces slight aliasing artifacts (as seen in the first row of Fig.\ref{fig:vae-comparison}), which can be mitigated by introducing eikonal regularization.

\begin{table}[t]
\begin{center}
\caption{The ablation for data quality and quantity.}
\vspace{1em}
\label{tab: data_abaltion}
\begin{sc}
\resizebox{0.48\textwidth}{!}{
\begin{tabular}{| c| c|c|c|}
\hline
Dataset  & Size & Data-building System &Normal-FID $\downarrow$\\
\hline
\multirow{2}{*}{Objaverse}  &$800K$ & \ding{55}& 11.61\\
\cline{2-4}
  &$180K$ & \ding{51}& 9.47 \\
\hline
\method{}  &$2M$ & \ding{51} & 5.81\\
\hline
\end{tabular}
}
\end{sc}
\end{center}
\vspace{-2em}
\end{table}
\subsection{Ablation for Data-Building System}
To demonstrate the effectiveness of the data-building system proposed by \method{}, we implement ablation experiments on both data quality and quantity. Using the optimal R-Flow training settings (first row of Tab.\ref{tab: diffusion_improvements_abaltion}), we replaced the $180K$ Objaverse dataset produced by \method{} with the original $800K$ Objaverse dataset, which had not undergone scoring, filtering, orientation fixing, untextured model processing, or internal processing of the converted watertight model. This experiment demonstrates the effect of data quality. Similarly, under the same R-Flow settings, we expanded the high-quality dataset from $180K$ Objaverse to $2M$ \method{} to evaluate the effect of data quantity.

As shown in the first two rows of Tab.\ref{tab: data_abaltion}, although our data-building system reduced the $800K$ Objaverse dataset to $180K$, the improved data quality resulted in better generation results, demonstrating that, when training with in-the-wild data, quality outweighs quantity. Furthermore, as seen in last two rows of Tab.\ref{tab: data_abaltion}, increasing the high-quality dataset from $180K$ to $2M$ led to a significant boost in generation performance, showing that with high-quality data, scaling up data size is crucial for achieving better results. 
Additionally, the overall quantitative results in the Tab.\ref{tab: data_abaltion} show that the performance improvement gained from $2M$ high-quality data size is greater than that from improving data quality alone. Furthermore, after enhancing data quality, performance continues to improve with an increase in data size, without encountering a bottleneck at the current training scale.

\subsection{The Visualization for Flow Model Ablation}
\begin{figure*}
    \centering
    \includegraphics[width=\linewidth]{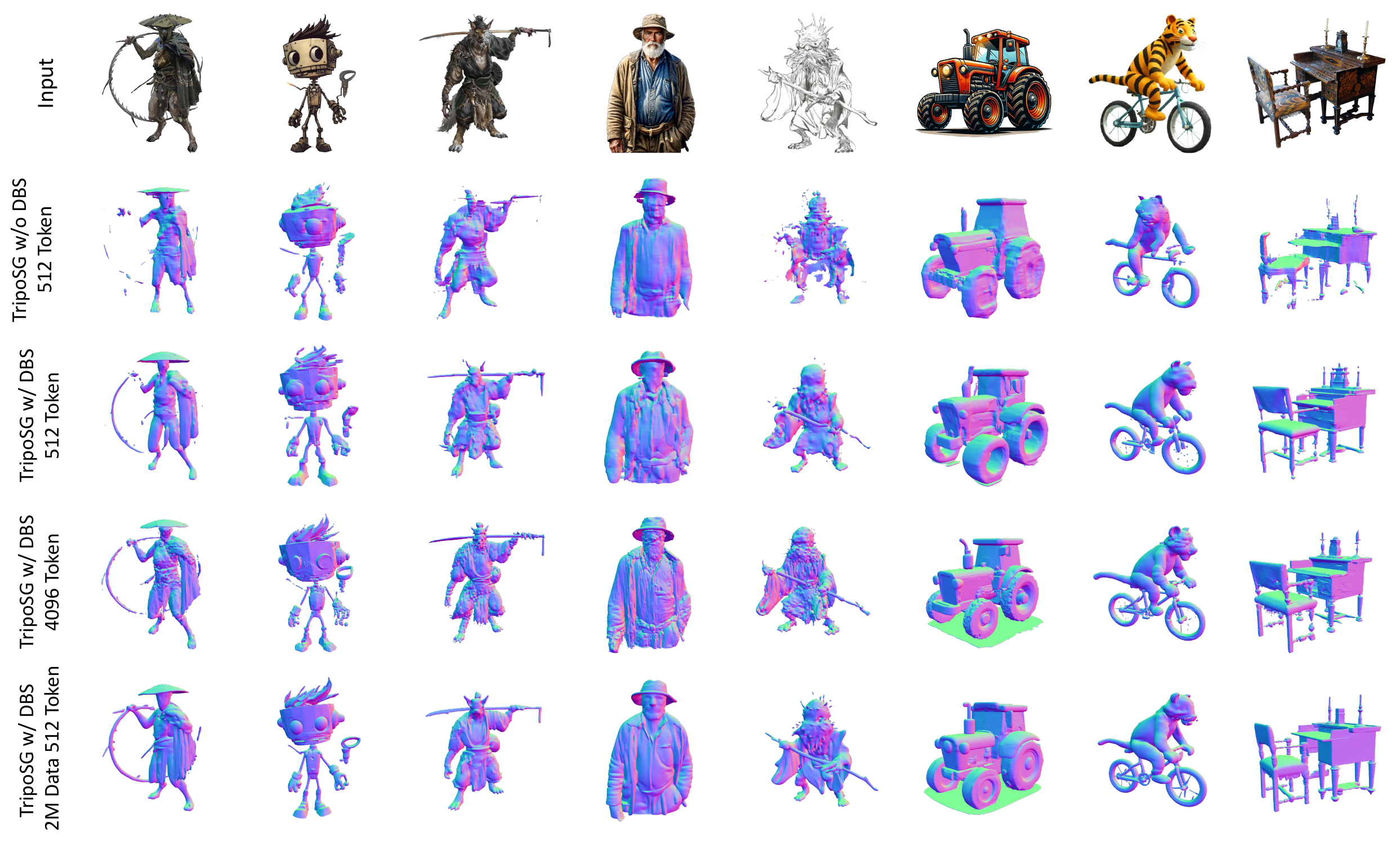}
    \vspace{-1em}
    \caption{Visualization results of the flow model ablation experiments. `DBS' is the abbreviation for Data-Building System.
    }
    \vspace{-1em}
    \label{fig:ablation_vis}
\end{figure*}

In addition to the quantitative results, we also performed a visualization analysis of the core experiments, as shown in Fig.\ref{fig:ablation_vis}. The rows 1, 2, 4 correspond to the three experimental results in Tab.\ref{tab: data_abaltion}, while the row 3 corresponds to the row 4 of results in Tab.\ref{tab: diffusion_scaling_abaltion}. The visualization reveals several insights consistent with the quantitative results: (1) Data quality is more important than the size of raw in-the-wild data (row 1 vs. row 2). (2) The improvement from increasing high-quality data size is more obvious than the improvement from resolution (row 2 vs. row 3 vs. row 4). (3) Increasing the size of high-quality data ($2M$) provides a greater boost to performance than merely improving data quality. After improving data quality, performance continues to improve with increased data size without encountering bottlenecks at the current training scale.

\section{Conclusion and Discussion}

\subsection{Conclusion}
We present \method{}, a new image-to-3D generation model via the rectified-flow-based transformer. To efficiently train the model for high-fidelity shape generation, we propose a data-building system to process data from original datasets.
Compared to using all in-the-wild 3D models in the training dataset, filtered and fixed high-quality data can be properly reproduced into training data and effectively improve the model's training performance.
Additionally, we leverage the advance of SDF representation with surface normal guidance and eikonal regularization for finer geometry details and avoid aliasing artifacts.
Furthermore, a rectified-flow-based transformer with MoE and a high-resolution strategy is introduced for the scale-up training.
Experiments demonstrate that \method{} can generate high-fidelity 3D shapes, leading to a new state-of-the-art performance.

\subsection{Discussion}
In recent years, 3D generation has followed a unique exploration route, with methods such as using text-to-image models as priors for 3D generation via the SDS solution, such as DreamFusion\cite{poole2022dreamfusion}, and leveraging decoder-only transformer architectures to reconstruct 3D models from single or multiple views, such as LRM\cite{hong2023lrm}. However, due to the scarcity of large-scale datasets and limited experience in scaling up training for 3D generation tasks, the large-scale flow models, which have proven highly successful in 2D image and video generation, have not been widely applied to 3D generation. \method{} has deeply explored the 3D flow route from the perspective of data and training, successfully achieving 3D generation with strong generalization, exceptional detail, and high fidelity. It has effectively replicated the success of image and video generation architectures in the field of 3D generation. Through \method{}, 3D generation now aligns with image and video generation in terms of architecture and development stage, allowing the field of 3D generation to draw upon the wealth of architectures and training experience from 2D image and video generation.

Looking ahead, we can further scale up model parameters and training data, and employ more fine-grained conditional information injection methods to generate even more detailed 3D models. Additionally, based on the \method{} foundation, we can also explore tasks such as 3D model super-resolution, scene generation, and stylization.

\bibliography{main}

\begin{thebibliography}{102}
\providecommand{\natexlab}[1]{#1}
\providecommand{\url}[1]{\texttt{#1}}
\expandafter\ifx\csname urlstyle\endcsname\relax
  \providecommand{\doi}[1]{doi: #1}\else
  \providecommand{\doi}{doi: \begingroup \urlstyle{rm}\Url}\fi

\bibitem[Bain et~al.(2021)Bain, Nagrani, Varol, and Zisserman]{bain2021frozen}
Bain, M., Nagrani, A., Varol, G., and Zisserman, A.
\newblock Frozen in time: A joint video and image encoder for end-to-end retrieval.
\newblock In \emph{Proceedings of the IEEE/CVF international conference on computer vision}, pp.\  1728--1738, 2021.

\bibitem[Bao et~al.(2023)Bao, Nie, Xue, Cao, Li, Su, and Zhu]{bao2023all}
Bao, F., Nie, S., Xue, K., Cao, Y., Li, C., Su, H., and Zhu, J.
\newblock All are worth words: A vit backbone for diffusion models.
\newblock In \emph{Proceedings of the IEEE/CVF conference on computer vision and pattern recognition}, pp.\  22669--22679, 2023.

\bibitem[Bensadoun et~al.(2024)Bensadoun, Kleiman, Azuri, Harosh, Vedaldi, Neverova, and Gafni]{bensadoun2024meta}
Bensadoun, R., Kleiman, Y., Azuri, I., Harosh, O., Vedaldi, A., Neverova, N., and Gafni, O.
\newblock Meta 3d texturegen: fast and consistent texture generation for 3d objects.
\newblock \emph{arXiv preprint arXiv:2407.02430}, 2024.

\bibitem[blackforestlabs(2024)]{flux}
blackforestlabs.
\newblock Flux.
\newblock https://github.com/black-forest-labs/flux, 2024.

\bibitem[Brooks et~al.(2024)Brooks, Peebles, Holmes, DePue, Guo, Jing, Schnurr, Taylor, Luhman, Luhman, Ng, Wang, and Ramesh]{videoworldsimulators2024}
Brooks, T., Peebles, B., Holmes, C., DePue, W., Guo, Y., Jing, L., Schnurr, D., Taylor, J., Luhman, T., Luhman, E., Ng, C., Wang, R., and Ramesh, A.
\newblock Video generation models as world simulators.
\newblock 2024.
\newblock URL \url{https://openai.com/research/video-generation-models-as-world-simulators}.

\bibitem[Chan et~al.(2023)Chan, Nagano, Chan, Bergman, Park, Levy, Aittala, Mello, Karras, and Wetzstein]{DBLP:conf/iccv/ChanNCBPLAMKW23}
Chan, E.~R., Nagano, K., Chan, M.~A., Bergman, A.~W., Park, J.~J., Levy, A., Aittala, M., Mello, S.~D., Karras, T., and Wetzstein, G.
\newblock Generative novel view synthesis with 3d-aware diffusion models.
\newblock In \emph{{IEEE/CVF} {ICCV}}, pp.\  4194--4206, 2023.

\bibitem[Chang et~al.(2015)Chang, Funkhouser, Guibas, Hanrahan, Huang, Li, Savarese, Savva, Song, Su, et~al.]{chang2015shapenet}
Chang, A.~X., Funkhouser, T., Guibas, L., Hanrahan, P., Huang, Q., Li, Z., Savarese, S., Savva, M., Song, S., Su, H., et~al.
\newblock Shapenet: An information-rich 3d model repository.
\newblock \emph{arXiv preprint arXiv:1512.03012}, 2015.

\bibitem[Chen et~al.(2023)Chen, Chen, Jiao, and Jia]{DBLP:conf/iccv/ChenCJJ23}
Chen, R., Chen, Y., Jiao, N., and Jia, K.
\newblock Fantasia3d: Disentangling geometry and appearance for high-quality text-to-3d content creation.
\newblock In \emph{{IEEE/CVF} {ICCV}}, pp.\  22189--22199, 2023.

\bibitem[Cheng et~al.(2023)Cheng, Lee, Tulyakov, Schwing, and Gui]{DBLP:conf/cvpr/ChengLTSG23}
Cheng, Y., Lee, H., Tulyakov, S., Schwing, A.~G., and Gui, L.
\newblock Sdfusion: Multimodal 3d shape completion, reconstruction, and generation.
\newblock In \emph{{IEEE/CVF} {CVPR}}, pp.\  4456--4465, 2023.

\bibitem[Dehghani et~al.(2023)Dehghani, Djolonga, Mustafa, Padlewski, Heek, Gilmer, Steiner, Caron, Geirhos, Alabdulmohsin, et~al.]{dehghani2023scaling}
Dehghani, M., Djolonga, J., Mustafa, B., Padlewski, P., Heek, J., Gilmer, J., Steiner, A.~P., Caron, M., Geirhos, R., Alabdulmohsin, I., et~al.
\newblock Scaling vision transformers to 22 billion parameters.
\newblock In \emph{International Conference on Machine Learning}, pp.\  7480--7512. PMLR, 2023.

\bibitem[Deitke et~al.(2023)Deitke, Schwenk, Salvador, Weihs, Michel, VanderBilt, Schmidt, Ehsani, Kembhavi, and Farhadi]{deitke2023objaverse}
Deitke, M., Schwenk, D., Salvador, J., Weihs, L., Michel, O., VanderBilt, E., Schmidt, L., Ehsani, K., Kembhavi, A., and Farhadi, A.
\newblock Objaverse: A universe of annotated 3d objects.
\newblock In \emph{Proceedings of the IEEE/CVF Conference on Computer Vision and Pattern Recognition}, pp.\  13142--13153, 2023.

\bibitem[Deitke et~al.(2024)Deitke, Liu, Wallingford, Ngo, Michel, Kusupati, Fan, Laforte, Voleti, Gadre, et~al.]{deitke2024objaverse}
Deitke, M., Liu, R., Wallingford, M., Ngo, H., Michel, O., Kusupati, A., Fan, A., Laforte, C., Voleti, V., Gadre, S.~Y., et~al.
\newblock Objaverse-xl: A universe of 10m+ 3d objects.
\newblock \emph{Advances in Neural Information Processing Systems}, 36, 2024.

\bibitem[Deng et~al.(2009)Deng, Dong, Socher, Li, Li, and Fei-Fei]{deng2009imagenet}
Deng, J., Dong, W., Socher, R., Li, L.-J., Li, K., and Fei-Fei, L.
\newblock Imagenet: A large-scale hierarchical image database.
\newblock In \emph{2009 IEEE conference on computer vision and pattern recognition}, pp.\  248--255. Ieee, 2009.

\bibitem[Esser et~al.(2024)Esser, Kulal, Blattmann, Entezari, M{\"u}ller, Saini, Levi, Lorenz, Sauer, Boesel, et~al.]{esser2024scaling}
Esser, P., Kulal, S., Blattmann, A., Entezari, R., M{\"u}ller, J., Saini, H., Levi, Y., Lorenz, D., Sauer, A., Boesel, F., et~al.
\newblock Scaling rectified flow transformers for high-resolution image synthesis.
\newblock In \emph{Forty-first International Conference on Machine Learning}, 2024.

\bibitem[Fan et~al.(2017)Fan, Su, and Guibas]{DBLP:conf/cvpr/FanSG17}
Fan, H., Su, H., and Guibas, L.~J.
\newblock A point set generation network for 3d object reconstruction from a single image.
\newblock In \emph{{IEEE/CVF} {CVPR}}, pp.\  2463--2471, 2017.

\bibitem[Fei et~al.(2024)Fei, Fan, Yu, Li, and Huang]{fei2024scaling}
Fei, Z., Fan, M., Yu, C., Li, D., and Huang, J.
\newblock Scaling diffusion transformers to 16 billion parameters.
\newblock \emph{arXiv preprint arXiv:2407.11633}, 2024.

\bibitem[Girdhar et~al.(2016)Girdhar, Fouhey, Rodriguez, and Gupta]{DBLP:conf/eccv/GirdharFRG16}
Girdhar, R., Fouhey, D.~F., Rodriguez, M., and Gupta, A.
\newblock Learning a predictable and generative vector representation for objects.
\newblock In Leibe, B., Matas, J., Sebe, N., and Welling, M. (eds.), \emph{{ECCV}}, volume 9910, pp.\  484--499, 2016.

\bibitem[Heusel et~al.(2017)Heusel, Ramsauer, Unterthiner, Nessler, and Hochreiter]{heusel2017gans}
Heusel, M., Ramsauer, H., Unterthiner, T., Nessler, B., and Hochreiter, S.
\newblock Gans trained by a two time-scale update rule converge to a local nash equilibrium.
\newblock \emph{Advances in neural information processing systems}, 30, 2017.

\bibitem[Ho et~al.(2020)Ho, Jain, and Abbeel]{DBLP:conf/nips/HoJA20}
Ho, J., Jain, A., and Abbeel, P.
\newblock Denoising diffusion probabilistic models.
\newblock In Larochelle, H., Ranzato, M., Hadsell, R., Balcan, M., and Lin, H. (eds.), \emph{NeurIPS}, 2020.

\bibitem[Hong et~al.(2024)Hong, Tang, Cao, Shi, Wu, Chen, Wang, Pan, Lin, and Liu]{DBLP:journals/corr/abs-2403-02234}
Hong, F., Tang, J., Cao, Z., Shi, M., Wu, T., Chen, Z., Wang, T., Pan, L., Lin, D., and Liu, Z.
\newblock 3dtopia: Large text-to-3d generation model with hybrid diffusion priors.
\newblock \emph{CoRR}, abs/2403.02234, 2024.

\bibitem[Hong et~al.(2023)Hong, Zhang, Gu, Bi, Zhou, Liu, Liu, Sunkavalli, Bui, and Tan]{hong2023lrm}
Hong, Y., Zhang, K., Gu, J., Bi, S., Zhou, Y., Liu, D., Liu, F., Sunkavalli, K., Bui, T., and Tan, H.
\newblock Lrm: Large reconstruction model for single image to 3d.
\newblock \emph{arXiv preprint arXiv:2311.04400}, 2023.

\bibitem[Huang et~al.(2018)Huang, Su, and Guibas]{DBLP:journals/corr/abs-1802-01698}
Huang, J., Su, H., and Guibas, L.~J.
\newblock Robust watertight manifold surface generation method for shapenet models.
\newblock \emph{CoRR}, abs/1802.01698, 2018.

\bibitem[Huang et~al.(2020)Huang, Zhou, and Guibas]{DBLP:journals/corr/abs-2005-11621}
Huang, J., Zhou, Y., and Guibas, L.~J.
\newblock Manifoldplus: {A} robust and scalable watertight manifold surface generation method for triangle soups.
\newblock \emph{CoRR}, abs/2005.11621, 2020.

\bibitem[Hui et~al.(2022)Hui, Li, Hu, and Fu]{DBLP:conf/siggrapha/HuiLHF22}
Hui, K., Li, R., Hu, J., and Fu, C.
\newblock Neural wavelet-domain diffusion for 3d shape generation.
\newblock In Jung, S.~K., Lee, J., and Bargteil, A.~W. (eds.), \emph{{SIGGRAPH} Asia}, pp.\  24:1--24:9. {ACM}, 2022.

\bibitem[Jun \& Nichol(2023)Jun and Nichol]{DBLP:journals/corr/abs-2305-02463}
Jun, H. and Nichol, A.
\newblock Shap-e: Generating conditional 3d implicit functions.
\newblock \emph{CoRR}, abs/2305.02463, 2023.

\bibitem[Lan et~al.(2024)Lan, Hong, Yang, Zhou, Meng, Dai, Pan, and Loy]{lan2024ln3diff}
Lan, Y., Hong, F., Yang, S., Zhou, S., Meng, X., Dai, B., Pan, X., and Loy, C.~C.
\newblock Ln3diff: Scalable latent neural fields diffusion for speedy 3d generation.
\newblock In \emph{ECCV}, 2024.

\bibitem[Li et~al.(2023)Li, Tan, Zhang, Xu, Luan, Xu, Hong, Sunkavalli, Shakhnarovich, and Bi]{li2023instant3d}
Li, J., Tan, H., Zhang, K., Xu, Z., Luan, F., Xu, Y., Hong, Y., Sunkavalli, K., Shakhnarovich, G., and Bi, S.
\newblock Instant3d: Fast text-to-3d with sparse-view generation and large reconstruction model.
\newblock \emph{arXiv preprint arXiv:2311.06214}, 2023.

\bibitem[Li et~al.(2024{\natexlab{a}})Li, Yang, Kuang, Wu, Wang, Xiao, and Chen]{li2024controlnet++}
Li, M., Yang, T., Kuang, H., Wu, J., Wang, Z., Xiao, X., and Chen, C.
\newblock Controlnet++: Improving conditional controls with efficient consistency feedback.
\newblock \emph{arXiv preprint arXiv:2404.07987}, 2024{\natexlab{a}}.

\bibitem[Li et~al.(2024{\natexlab{b}})Li, Liu, Long, Zhang, Lin, Li, Qi, Zhang, Luo, Tan, Wang, Liu, and Guo]{DBLP:journals/corr/abs-2405-11616}
Li, P., Liu, Y., Long, X., Zhang, F., Lin, C., Li, M., Qi, X., Zhang, S., Luo, W., Tan, P., Wang, W., Liu, Q., and Guo, Y.
\newblock Era3d: High-resolution multiview diffusion using efficient row-wise attention.
\newblock \emph{CoRR}, abs/2405.11616, 2024{\natexlab{b}}.

\bibitem[Li et~al.(2024{\natexlab{c}})Li, Liu, Chen, Liang, Chen, Tan, and Long]{li2024craftsman}
Li, W., Liu, J., Chen, R., Liang, Y., Chen, X., Tan, P., and Long, X.
\newblock Craftsman: High-fidelity mesh generation with 3d native generation and interactive geometry refiner.
\newblock \emph{arXiv preprint arXiv:2405.14979}, 2024{\natexlab{c}}.

\bibitem[Liang et~al.(2024)Liang, Yang, Lin, Li, Xu, and Chen]{liang2024luciddreamer}
Liang, Y., Yang, X., Lin, J., Li, H., Xu, X., and Chen, Y.
\newblock Luciddreamer: Towards high-fidelity text-to-3d generation via interval score matching.
\newblock In \emph{IEEE/CVF CVPR}, pp.\  6517--6526, 2024.

\bibitem[Lin et~al.(2023)Lin, Gao, Tang, Takikawa, Zeng, Huang, Kreis, Fidler, Liu, and Lin]{DBLP:conf/cvpr/Lin0TTZHKF0L23}
Lin, C., Gao, J., Tang, L., Takikawa, T., Zeng, X., Huang, X., Kreis, K., Fidler, S., Liu, M., and Lin, T.
\newblock Magic3d: High-resolution text-to-3d content creation.
\newblock In \emph{{IEEE} {CVPR}}, pp.\  300--309, 2023.

\bibitem[Liu et~al.(2023{\natexlab{a}})Liu, Xu, Jin, Chen, T., Xu, and Su]{DBLP:conf/nips/LiuXJCTXS23}
Liu, M., Xu, C., Jin, H., Chen, L., T., M.~V., Xu, Z., and Su, H.
\newblock One-2-3-45: Any single image to 3d mesh in 45 seconds without per-shape optimization.
\newblock In \emph{NeurIPS}, 2023{\natexlab{a}}.

\bibitem[Liu et~al.(2024{\natexlab{a}})Liu, Zeng, Wei, Shi, Chen, Xu, Zhang, Wang, Zhang, Liu, Wu, and Su]{DBLP:journals/corr/abs-2408-10198}
Liu, M., Zeng, C., Wei, X., Shi, R., Chen, L., Xu, C., Zhang, M., Wang, Z., Zhang, X., Liu, I., Wu, H., and Su, H.
\newblock Meshformer: High-quality mesh generation with 3d-guided reconstruction model.
\newblock \emph{CoRR}, abs/2408.10198, 2024{\natexlab{a}}.

\bibitem[Liu et~al.(2023{\natexlab{b}})Liu, Wu, Hoorick, Tokmakov, Zakharov, and Vondrick]{DBLP:conf/iccv/LiuWHTZV23}
Liu, R., Wu, R., Hoorick, B.~V., Tokmakov, P., Zakharov, S., and Vondrick, C.
\newblock Zero-1-to-3: Zero-shot one image to 3d object.
\newblock In \emph{{IEEE/CVF} {ICCV}}, pp.\  9264--9275, 2023{\natexlab{b}}.

\bibitem[Liu et~al.(2023{\natexlab{c}})Liu, Gong, and Liu]{DBLP:conf/iclr/LiuG023}
Liu, X., Gong, C., and Liu, Q.
\newblock Flow straight and fast: Learning to generate and transfer data with rectified flow.
\newblock In \emph{{ICLR}}, 2023{\natexlab{c}}.

\bibitem[Liu et~al.(2024{\natexlab{b}})Liu, Lin, Zeng, Long, Liu, Komura, and Wang]{DBLP:conf/iclr/LiuLZLLKW24}
Liu, Y., Lin, C., Zeng, Z., Long, X., Liu, L., Komura, T., and Wang, W.
\newblock Syncdreamer: Generating multiview-consistent images from a single-view image.
\newblock In \emph{{ICLR}}, 2024{\natexlab{b}}.

\bibitem[Liu et~al.(2023{\natexlab{d}})Liu, Li, Lin, Yu, Peng, Cao, Qi, Huang, Liang, and Ouyang]{liu2023unidream}
Liu, Z., Li, Y., Lin, Y., Yu, X., Peng, S., Cao, Y.-P., Qi, X., Huang, X., Liang, D., and Ouyang, W.
\newblock Unidream: Unifying diffusion priors for relightable text-to-3d generation.
\newblock \emph{arXiv preprint arXiv:2312.08754}, 2023{\natexlab{d}}.

\bibitem[Long et~al.(2024)Long, Guo, Lin, Liu, Dou, Liu, Ma, Zhang, Habermann, Theobalt, et~al.]{long2024wonder3d}
Long, X., Guo, Y.-C., Lin, C., Liu, Y., Dou, Z., Liu, L., Ma, Y., Zhang, S.-H., Habermann, M., Theobalt, C., et~al.
\newblock Wonder3d: Single image to 3d using cross-domain diffusion.
\newblock In \emph{IEEE/CVF CVPR}, pp.\  9970--9980, 2024.

\bibitem[Lorensen \& Cline(1987)Lorensen and Cline]{DBLP:conf/siggraph/LorensenC87}
Lorensen, W.~E. and Cline, H.~E.
\newblock Marching cubes: {A} high resolution 3d surface construction algorithm.
\newblock In Stone, M.~C. (ed.), \emph{Proceedings of the {SIGGRAPH}}, pp.\  163--169, 1987.

\bibitem[Melas{-}Kyriazi et~al.(2023)Melas{-}Kyriazi, Rupprecht, and Vedaldi]{DBLP:conf/cvpr/Melas-Kyriazi0V23}
Melas{-}Kyriazi, L., Rupprecht, C., and Vedaldi, A.
\newblock Pc\({}^{\mbox{2}}\): Projection-conditioned point cloud diffusion for single-image 3d reconstruction.
\newblock In \emph{{IEEE/CVF} {CVPR}}, pp.\  12923--12932, 2023.

\bibitem[Mescheder et~al.(2019)Mescheder, Oechsle, Niemeyer, Nowozin, and Geiger]{DBLP:conf/cvpr/MeschederONNG19}
Mescheder, L.~M., Oechsle, M., Niemeyer, M., Nowozin, S., and Geiger, A.
\newblock Occupancy networks: Learning 3d reconstruction in function space.
\newblock In \emph{{IEEE/CVF} {CVPR}}, pp.\  4460--4470, 2019.

\bibitem[Metzer et~al.(2023)Metzer, Richardson, Patashnik, Giryes, and Cohen{-}Or]{DBLP:conf/cvpr/MetzerRPGC23}
Metzer, G., Richardson, E., Patashnik, O., Giryes, R., and Cohen{-}Or, D.
\newblock Latent-nerf for shape-guided generation of 3d shapes and textures.
\newblock In \emph{{IEEE/CVF} {CVPR}}, pp.\  12663--12673, 2023.

\bibitem[Mildenhall et~al.(2020)Mildenhall, Srinivasan, Tancik, Barron, Ramamoorthi, and Ng]{DBLP:conf/eccv/MildenhallSTBRN20}
Mildenhall, B., Srinivasan, P.~P., Tancik, M., Barron, J.~T., Ramamoorthi, R., and Ng, R.
\newblock Nerf: Representing scenes as neural radiance fields for view synthesis.
\newblock In Vedaldi, A., Bischof, H., Brox, T., and Frahm, J. (eds.), \emph{ECCV}, volume 12346, pp.\  405--421, 2020.

\bibitem[M{\"{u}}ller et~al.(2023)M{\"{u}}ller, Siddiqui, Porzi, Bul{\`{o}}, Kontschieder, and Nie{\ss}ner]{DBLP:conf/cvpr/MullerSPBKN23}
M{\"{u}}ller, N., Siddiqui, Y., Porzi, L., Bul{\`{o}}, S.~R., Kontschieder, P., and Nie{\ss}ner, M.
\newblock Diffrf: Rendering-guided 3d radiance field diffusion.
\newblock In \emph{{IEEE/CVF} {CVPR}}, pp.\  4328--4338, 2023.

\bibitem[Newcombe et~al.(2011)Newcombe, Izadi, Hilliges, Molyneaux, Kim, Davison, Kohli, Shotton, Hodges, and Fitzgibbon]{DBLP:conf/ismar/NewcombeIHMKDKSHF11}
Newcombe, R.~A., Izadi, S., Hilliges, O., Molyneaux, D., Kim, D., Davison, A.~J., Kohli, P., Shotton, J., Hodges, S., and Fitzgibbon, A.~W.
\newblock Kinectfusion: Real-time dense surface mapping and tracking.
\newblock In \emph{{IEEE} {ISMAR}}, pp.\  127--136, 2011.

\bibitem[Nichol et~al.(2022)Nichol, Jun, Dhariwal, Mishkin, and Chen]{DBLP:journals/corr/abs-2212-08751}
Nichol, A., Jun, H., Dhariwal, P., Mishkin, P., and Chen, M.
\newblock Point-e: {A} system for generating 3d point clouds from complex prompts.
\newblock \emph{CoRR}, abs/2212.08751, 2022.

\bibitem[OpenAI(2023{\natexlab{a}})]{DBLP:journals/corr/abs-2303-08774}
OpenAI.
\newblock {GPT-4} technical report.
\newblock \emph{CoRR}, abs/2303.08774, 2023{\natexlab{a}}.

\bibitem[OpenAI(2023{\natexlab{b}})]{openai2023gpt4v}
OpenAI.
\newblock Gpt-4v(ision) system card.
\newblock \url{https://cdn.openai.com/papers/GPTV_System_Card.pdf}, 2023{\natexlab{b}}.
\newblock Accessed: 2023-09-25.

\bibitem[Oquab et~al.(2023)Oquab, Darcet, Moutakanni, Vo, Szafraniec, Khalidov, Fernandez, Haziza, Massa, El-Nouby, et~al.]{oquab2023dinov2}
Oquab, M., Darcet, T., Moutakanni, T., Vo, H., Szafraniec, M., Khalidov, V., Fernandez, P., Haziza, D., Massa, F., El-Nouby, A., et~al.
\newblock Dinov2: Learning robust visual features without supervision.
\newblock \emph{arXiv preprint arXiv:2304.07193}, 2023.

\bibitem[Peebles \& Xie(2023)Peebles and Xie]{peebles2023scalable}
Peebles, W. and Xie, S.
\newblock Scalable diffusion models with transformers.
\newblock In \emph{Proceedings of the IEEE/CVF International Conference on Computer Vision}, pp.\  4195--4205, 2023.

\bibitem[Poole et~al.(2022)Poole, Jain, Barron, and Mildenhall]{poole2022dreamfusion}
Poole, B., Jain, A., Barron, J.~T., and Mildenhall, B.
\newblock Dreamfusion: Text-to-3d using 2d diffusion.
\newblock \emph{arXiv preprint arXiv:2209.14988}, 2022.

\bibitem[Poole et~al.(2023)Poole, Jain, Barron, and Mildenhall]{DBLP:conf/iclr/PooleJBM23}
Poole, B., Jain, A., Barron, J.~T., and Mildenhall, B.
\newblock Dreamfusion: Text-to-3d using 2d diffusion.
\newblock In \emph{{ICLR}}, 2023.

\bibitem[Radford et~al.(2021)Radford, Kim, Hallacy, Ramesh, Goh, Agarwal, Sastry, Askell, Mishkin, Clark, et~al.]{radford2021learning}
Radford, A., Kim, J.~W., Hallacy, C., Ramesh, A., Goh, G., Agarwal, S., Sastry, G., Askell, A., Mishkin, P., Clark, J., et~al.
\newblock Learning transferable visual models from natural language supervision.
\newblock In \emph{International conference on machine learning}, pp.\  8748--8763. PMLR, 2021.

\bibitem[Ramesh et~al.(2021)Ramesh, Pavlov, Goh, Gray, Voss, Radford, Chen, and Sutskever]{DBLP:conf/icml/RameshPGGVRCS21}
Ramesh, A., Pavlov, M., Goh, G., Gray, S., Voss, C., Radford, A., Chen, M., and Sutskever, I.
\newblock Zero-shot text-to-image generation.
\newblock In \emph{{ICML}}, volume 139, pp.\  8821--8831, 2021.

\bibitem[Riquelme et~al.(2021)Riquelme, Puigcerver, Mustafa, Neumann, Jenatton, Susano~Pinto, Keysers, and Houlsby]{riquelme2021scaling}
Riquelme, C., Puigcerver, J., Mustafa, B., Neumann, M., Jenatton, R., Susano~Pinto, A., Keysers, D., and Houlsby, N.
\newblock Scaling vision with sparse mixture of experts.
\newblock \emph{Advances in Neural Information Processing Systems}, 34:\penalty0 8583--8595, 2021.

\bibitem[Rombach et~al.(2022)Rombach, Blattmann, Lorenz, Esser, and Ommer]{rombach2022high}
Rombach, R., Blattmann, A., Lorenz, D., Esser, P., and Ommer, B.
\newblock High-resolution image synthesis with latent diffusion models.
\newblock In \emph{Proceedings of the IEEE/CVF conference on computer vision and pattern recognition}, pp.\  10684--10695, 2022.

\bibitem[Schuhmann et~al.(2022)Schuhmann, Beaumont, Vencu, Gordon, Wightman, Cherti, Coombes, Katta, Mullis, Wortsman, et~al.]{schuhmann2022laion}
Schuhmann, C., Beaumont, R., Vencu, R., Gordon, C., Wightman, R., Cherti, M., Coombes, T., Katta, A., Mullis, C., Wortsman, M., et~al.
\newblock Laion-5b: An open large-scale dataset for training next generation image-text models.
\newblock \emph{Advances in Neural Information Processing Systems}, 35:\penalty0 25278--25294, 2022.

\bibitem[Shi et~al.(2023{\natexlab{a}})Shi, Chen, Zhang, Liu, Xu, Wei, Chen, Zeng, and Su]{DBLP:journals/corr/abs-2310-15110}
Shi, R., Chen, H., Zhang, Z., Liu, M., Xu, C., Wei, X., Chen, L., Zeng, C., and Su, H.
\newblock Zero123++: a single image to consistent multi-view diffusion base model.
\newblock \emph{CoRR}, abs/2310.15110, 2023{\natexlab{a}}.

\bibitem[Shi et~al.(2023{\natexlab{b}})Shi, Wang, Ye, Long, Li, and Yang]{shi2023mvdream}
Shi, Y., Wang, P., Ye, J., Long, M., Li, K., and Yang, X.
\newblock Mvdream: Multi-view diffusion for 3d generation.
\newblock \emph{arXiv preprint arXiv:2308.16512}, 2023{\natexlab{b}}.

\bibitem[Shi et~al.(2024)Shi, Wang, Ye, Mai, Li, and Yang]{DBLP:conf/iclr/ShiWYMLY24}
Shi, Y., Wang, P., Ye, J., Mai, L., Li, K., and Yang, X.
\newblock Mvdream: Multi-view diffusion for 3d generation.
\newblock In \emph{{ICLR}}, 2024.

\bibitem[Shue et~al.(2023)Shue, Chan, Po, Ankner, Wu, and Wetzstein]{DBLP:conf/cvpr/ShueCPA0W23}
Shue, J.~R., Chan, E.~R., Po, R., Ankner, Z., Wu, J., and Wetzstein, G.
\newblock 3d neural field generation using triplane diffusion.
\newblock In \emph{{IEEE/CVF} {CVPR}}, pp.\  20875--20886, 2023.

\bibitem[Singer et~al.(2023)Singer, Polyak, Hayes, Yin, An, Zhang, Hu, Yang, Ashual, Gafni, Parikh, Gupta, and Taigman]{DBLP:conf/iclr/SingerPH00ZHYAG23}
Singer, U., Polyak, A., Hayes, T., Yin, X., An, J., Zhang, S., Hu, Q., Yang, H., Ashual, O., Gafni, O., Parikh, D., Gupta, S., and Taigman, Y.
\newblock Make-a-video: Text-to-video generation without text-video data.
\newblock In \emph{{ICLR}}, 2023.

\bibitem[Tang et~al.(2023)Tang, Wang, Zhang, Zhang, Yi, Ma, and Chen]{DBLP:conf/iccv/TangWZZYM023}
Tang, J., Wang, T., Zhang, B., Zhang, T., Yi, R., Ma, L., and Chen, D.
\newblock Make-it-3d: High-fidelity 3d creation from {A} single image with diffusion prior.
\newblock In \emph{{IEEE} {ICCV}}, pp.\  22762--22772, 2023.

\bibitem[Tang et~al.(2024)Tang, Ren, Zhou, Liu, and Zeng]{DBLP:conf/iclr/TangRZ0Z24}
Tang, J., Ren, J., Zhou, H., Liu, Z., and Zeng, G.
\newblock Dreamgaussian: Generative gaussian splatting for efficient 3d content creation.
\newblock In \emph{{ICLR}}, 2024.

\bibitem[team~at Meta(2024)]{metamoviegen}
team~at Meta, T. M.~G.
\newblock Movie gen: A cast of media foundation models.
\newblock 2024.

\bibitem[Tochilkin et~al.(2024)Tochilkin, Pankratz, Liu, Huang, Letts, Li, Liang, Laforte, Jampani, and Cao]{tochilkin2024triposr}
Tochilkin, D., Pankratz, D., Liu, Z., Huang, Z., Letts, A., Li, Y., Liang, D., Laforte, C., Jampani, V., and Cao, Y.-P.
\newblock Triposr: Fast 3d object reconstruction from a single image.
\newblock \emph{arXiv preprint arXiv:2403.02151}, 2024.

\bibitem[Vaswani et~al.(2017)Vaswani, Shazeer, Parmar, Uszkoreit, Jones, Gomez, Kaiser, and Polosukhin]{DBLP:conf/nips/VaswaniSPUJGKP17}
Vaswani, A., Shazeer, N., Parmar, N., Uszkoreit, J., Jones, L., Gomez, A.~N., Kaiser, L., and Polosukhin, I.
\newblock Attention is all you need.
\newblock In \emph{NeurIPS}, pp.\  5998--6008, 2017.

\bibitem[von Platen et~al.(2022)von Platen, Patil, Lozhkov, Cuenca, Lambert, Rasul, Davaadorj, Nair, Paul, Berman, Xu, Liu, and Wolf]{von-platen-etal-2022-diffusers}
von Platen, P., Patil, S., Lozhkov, A., Cuenca, P., Lambert, N., Rasul, K., Davaadorj, M., Nair, D., Paul, S., Berman, W., Xu, Y., Liu, S., and Wolf, T.
\newblock Diffusers: State-of-the-art diffusion models.
\newblock \url{https://github.com/huggingface/diffusers}, 2022.

\bibitem[Wang et~al.(2023{\natexlab{a}})Wang, Du, Li, Yeh, and Shakhnarovich]{DBLP:conf/cvpr/WangDLYS23}
Wang, H., Du, X., Li, J., Yeh, R.~A., and Shakhnarovich, G.
\newblock Score jacobian chaining: Lifting pretrained 2d diffusion models for 3d generation.
\newblock In \emph{{IEEE} {CVPR}}, pp.\  12619--12629, 2023{\natexlab{a}}.

\bibitem[Wang et~al.(2018)Wang, Zhang, Li, Fu, Liu, and Jiang]{DBLP:conf/eccv/WangZLFLJ18}
Wang, N., Zhang, Y., Li, Z., Fu, Y., Liu, W., and Jiang, Y.
\newblock Pixel2mesh: Generating 3d mesh models from single {RGB} images.
\newblock In Ferrari, V., Hebert, M., Sminchisescu, C., and Weiss, Y. (eds.), \emph{{ECCV}}, volume 11215, pp.\  55--71, 2018.

\bibitem[Wang \& Shi(2023{\natexlab{a}})Wang and Shi]{DBLP:journals/corr/abs-2312-02201}
Wang, P. and Shi, Y.
\newblock Imagedream: Image-prompt multi-view diffusion for 3d generation.
\newblock \emph{CoRR}, abs/2312.02201, 2023{\natexlab{a}}.

\bibitem[Wang \& Shi(2023{\natexlab{b}})Wang and Shi]{wang2023imagedream}
Wang, P. and Shi, Y.
\newblock Imagedream: Image-prompt multi-view diffusion for 3d generation.
\newblock \emph{arXiv preprint arXiv:2312.02201}, 2023{\natexlab{b}}.

\bibitem[Wang et~al.(2022)Wang, Liu, and Tong]{DBLP:journals/tog/WangLT22}
Wang, P., Liu, Y., and Tong, X.
\newblock Dual octree graph networks for learning adaptive volumetric shape representations.
\newblock \emph{{ACM} Transactions on Graphics (TOG)}, 41\penalty0 (4):\penalty0 103:1--103:15, 2022.

\bibitem[Wang et~al.(2023{\natexlab{b}})Wang, Tan, Bi, Xu, Luan, Sunkavalli, Wang, Xu, and Zhang]{wang2023pf}
Wang, P., Tan, H., Bi, S., Xu, Y., Luan, F., Sunkavalli, K., Wang, W., Xu, Z., and Zhang, K.
\newblock Pf-lrm: Pose-free large reconstruction model for joint pose and shape prediction.
\newblock \emph{arXiv preprint arXiv:2311.12024}, 2023{\natexlab{b}}.

\bibitem[Wang et~al.(2023{\natexlab{c}})Wang, He, Li, Li, Yu, Ma, Li, Chen, Chen, Wang, et~al.]{wang2023internvid}
Wang, Y., He, Y., Li, Y., Li, K., Yu, J., Ma, X., Li, X., Chen, G., Chen, X., Wang, Y., et~al.
\newblock Internvid: A large-scale video-text dataset for multimodal understanding and generation.
\newblock \emph{arXiv preprint arXiv:2307.06942}, 2023{\natexlab{c}}.

\bibitem[Wang et~al.(2023{\natexlab{d}})Wang, Lu, Wang, Bao, Li, Su, and Zhu]{DBLP:conf/nips/Wang00BL0023}
Wang, Z., Lu, C., Wang, Y., Bao, F., Li, C., Su, H., and Zhu, J.
\newblock Prolificdreamer: High-fidelity and diverse text-to-3d generation with variational score distillation.
\newblock In \emph{NeurIPS}, 2023{\natexlab{d}}.

\bibitem[Wang et~al.(2024)Wang, Wang, Chen, Xiang, Chen, Yu, Li, Su, and Zhu]{wang2024crm}
Wang, Z., Wang, Y., Chen, Y., Xiang, C., Chen, S., Yu, D., Li, C., Su, H., and Zhu, J.
\newblock Crm: Single image to 3d textured mesh with convolutional reconstruction model.
\newblock \emph{arXiv preprint arXiv:2403.05034}, 2024.

\bibitem[Wei et~al.(2024)Wei, Zhang, Bi, Tan, Luan, Deschaintre, Sunkavalli, Su, and Xu]{wei2024meshlrm}
Wei, X., Zhang, K., Bi, S., Tan, H., Luan, F., Deschaintre, V., Sunkavalli, K., Su, H., and Xu, Z.
\newblock Meshlrm: Large reconstruction model for high-quality mesh.
\newblock \emph{arXiv preprint arXiv:2404.12385}, 2024.

\bibitem[Worchel et~al.(2022)Worchel, Diaz, Hu, Schreer, Feldmann, and Eisert]{DBLP:conf/cvpr/WorchelDHSFE22}
Worchel, M., Diaz, R., Hu, W., Schreer, O., Feldmann, I., and Eisert, P.
\newblock Multi-view mesh reconstruction with neural deferred shading.
\newblock In \emph{{IEEE/CVF} {CVPR}}, pp.\  6177--6187, 2022.

\bibitem[Wu et~al.(2017)Wu, Wang, Xue, Sun, Freeman, and Tenenbaum]{DBLP:conf/nips/0001WXSFT17}
Wu, J., Wang, Y., Xue, T., Sun, X., Freeman, B., and Tenenbaum, J.
\newblock Marrnet: 3d shape reconstruction via 2.5d sketches.
\newblock In \emph{NeurIPS}, pp.\  540--550, 2017.

\bibitem[Wu et~al.(2023)Wu, Ge, Wang, Lei, Gu, Shi, Hsu, Shan, Qie, and Shou]{DBLP:conf/iccv/WuGWLGSHSQS23}
Wu, J.~Z., Ge, Y., Wang, X., Lei, S.~W., Gu, Y., Shi, Y., Hsu, W., Shan, Y., Qie, X., and Shou, M.~Z.
\newblock Tune-a-video: One-shot tuning of image diffusion models for text-to-video generation.
\newblock In \emph{{IEEE/CVF} {ICCV}}, pp.\  7589--7599, 2023.

\bibitem[Wu et~al.(2024{\natexlab{a}})Wu, Liu, Cai, Yan, Wang, Hu, Duan, and Ma]{DBLP:journals/corr/abs-2405-20343}
Wu, K., Liu, F., Cai, Z., Yan, R., Wang, H., Hu, Y., Duan, Y., and Ma, K.
\newblock Unique3d: High-quality and efficient 3d mesh generation from a single image.
\newblock \emph{CoRR}, abs/2405.20343, 2024{\natexlab{a}}.

\bibitem[Wu et~al.(2020)Wu, Zhuang, Xu, Zhang, and Chen]{DBLP:conf/cvpr/WuZXZC20}
Wu, R., Zhuang, Y., Xu, K., Zhang, H., and Chen, B.
\newblock {PQ-NET:} {A} generative part seq2seq network for 3d shapes.
\newblock In \emph{{IEEE/CVF} {CVPR}}, pp.\  826--835, 2020.

\bibitem[Wu et~al.(2024{\natexlab{b}})Wu, Lin, Zhang, Zeng, Xu, Torr, Cao, and Yao]{wu2024direct3d}
Wu, S., Lin, Y., Zhang, F., Zeng, Y., Xu, J., Torr, P., Cao, X., and Yao, Y.
\newblock Direct3d: Scalable image-to-3d generation via 3d latent diffusion transformer.
\newblock \emph{arXiv preprint arXiv:2405.14832}, 2024{\natexlab{b}}.

\bibitem[Wu et~al.(2024{\natexlab{c}})Wu, Yang, Li, Zhang, Liu, Guibas, Lin, and Wetzstein]{DBLP:conf/cvpr/WuYLZLGLW24}
Wu, T., Yang, G., Li, Z., Zhang, K., Liu, Z., Guibas, L.~J., Lin, D., and Wetzstein, G.
\newblock Gpt-4v(ision) is a human-aligned evaluator for text-to-3d generation.
\newblock In \emph{{IEEE/CVF} {CVPR}}, pp.\  22227--22238, 2024{\natexlab{c}}.

\bibitem[Xu et~al.(2024)Xu, Cheng, Gao, Wang, Gao, and Shan]{xu2024instantmesh}
Xu, J., Cheng, W., Gao, Y., Wang, X., Gao, S., and Shan, Y.
\newblock Instantmesh: Efficient 3d mesh generation from a single image with sparse-view large reconstruction models.
\newblock \emph{arXiv preprint arXiv:2404.07191}, 2024.

\bibitem[Xu et~al.(2019)Xu, Wang, Ceylan, Mech, and Neumann]{DBLP:conf/nips/XuWCMN19}
Xu, Q., Wang, W., Ceylan, D., Mech, R., and Neumann, U.
\newblock {DISN:} deep implicit surface network for high-quality single-view 3d reconstruction.
\newblock In \emph{NeurIPS}, pp.\  490--500, 2019.

\bibitem[Xu et~al.(2023)Xu, Tan, Luan, Bi, Wang, Li, Shi, Sunkavalli, Wetzstein, Xu, et~al.]{xu2023dmv3d}
Xu, Y., Tan, H., Luan, F., Bi, S., Wang, P., Li, J., Shi, Z., Sunkavalli, K., Wetzstein, G., Xu, Z., et~al.
\newblock Dmv3d: Denoising multi-view diffusion using 3d large reconstruction model.
\newblock \emph{arXiv preprint arXiv:2311.09217}, 2023.

\bibitem[Yi et~al.(2024)Yi, Fang, Wang, Wu, Xie, Zhang, Liu, Tian, and Wang]{yi2024gaussiandreamer}
Yi, T., Fang, J., Wang, J., Wu, G., Xie, L., Zhang, X., Liu, W., Tian, Q., and Wang, X.
\newblock Gaussiandreamer: Fast generation from text to 3d gaussians by bridging 2d and 3d diffusion models.
\newblock In \emph{IEEE CVPR}, pp.\  6796--6807, 2024.

\bibitem[Yu et~al.(2021)Yu, Ye, Tancik, and Kanazawa]{DBLP:conf/cvpr/YuYTK21}
Yu, A., Ye, V., Tancik, M., and Kanazawa, A.
\newblock pixelnerf: Neural radiance fields from one or few images.
\newblock In \emph{{IEEE/CVF} {CVPR}}, pp.\  4578--4587, 2021.

\bibitem[Zeiler \& Fergus(2014)Zeiler and Fergus]{zeiler2014visualizing}
Zeiler, M.~D. and Fergus, R.
\newblock Visualizing and understanding convolutional networks.
\newblock In \emph{Computer Vision--ECCV 2014: 13th European Conference, Zurich, Switzerland, September 6-12, 2014, Proceedings, Part I 13}, pp.\  818--833. Springer, 2014.

\bibitem[Zeng et~al.(2022)Zeng, Vahdat, Williams, Gojcic, Litany, Fidler, and Kreis]{DBLP:conf/nips/zengVWGLFK22}
Zeng, X., Vahdat, A., Williams, F., Gojcic, Z., Litany, O., Fidler, S., and Kreis, K.
\newblock {LION:} latent point diffusion models for 3d shape generation.
\newblock In \emph{NeurIPS}, 2022.

\bibitem[Zhang \& Sennrich(2019)Zhang and Sennrich]{zhang2019root}
Zhang, B. and Sennrich, R.
\newblock Root mean square layer normalization.
\newblock \emph{Advances in Neural Information Processing Systems}, 32, 2019.

\bibitem[Zhang et~al.(2023)Zhang, Tang, Niessner, and Wonka]{zhang20233dshape2vecset}
Zhang, B., Tang, J., Niessner, M., and Wonka, P.
\newblock 3dshape2vecset: A 3d shape representation for neural fields and generative diffusion models.
\newblock \emph{ACM Transactions on Graphics (TOG)}, 42\penalty0 (4):\penalty0 1--16, 2023.

\bibitem[Zhang et~al.(2024{\natexlab{a}})Zhang, Bi, Tan, Xiangli, Zhao, Sunkavalli, and Xu]{zhang2024gs}
Zhang, K., Bi, S., Tan, H., Xiangli, Y., Zhao, N., Sunkavalli, K., and Xu, Z.
\newblock Gs-lrm: Large reconstruction model for 3d gaussian splatting.
\newblock \emph{arXiv preprint arXiv:2404.19702}, 2024{\natexlab{a}}.

\bibitem[Zhang et~al.(2024{\natexlab{b}})Zhang, Wang, Zhang, Qiu, Pang, Jiang, Yang, Xu, and Yu]{zhang2024clay}
Zhang, L., Wang, Z., Zhang, Q., Qiu, Q., Pang, A., Jiang, H., Yang, W., Xu, L., and Yu, J.
\newblock Clay: A controllable large-scale generative model for creating high-quality 3d assets.
\newblock \emph{ACM Transactions on Graphics (TOG)}, 43\penalty0 (4):\penalty0 1--20, 2024{\natexlab{b}}.

\bibitem[Zhao et~al.(2024)Zhao, Liu, Chen, Zeng, Wang, Cheng, Fu, Chen, Yu, and Gao]{zhao2024michelangelo}
Zhao, Z., Liu, W., Chen, X., Zeng, X., Wang, R., Cheng, P., Fu, B., Chen, T., Yu, G., and Gao, S.
\newblock Michelangelo: Conditional 3d shape generation based on shape-image-text aligned latent representation.
\newblock \emph{Advances in Neural Information Processing Systems}, 36, 2024.

\bibitem[Zheng et~al.(2023)Zheng, Pan, Wang, Tong, Liu, and Shum]{DBLP:journals/tog/ZhengPWTLS23}
Zheng, X., Pan, H., Wang, P., Tong, X., Liu, Y., and Shum, H.
\newblock Locally attentional {SDF} diffusion for controllable 3d shape generation.
\newblock \emph{{ACM} Trans. Graph.}, 42\penalty0 (4):\penalty0 91:1--91:13, 2023.

\bibitem[Zhou et~al.(2021)Zhou, Du, and Wu]{DBLP:conf/iccv/ZhouD021}
Zhou, L., Du, Y., and Wu, J.
\newblock 3d shape generation and completion through point-voxel diffusion.
\newblock In \emph{{IEEE/CVF} {ICCV}}, pp.\  5806--5815, 2021.

\bibitem[Zou et~al.(2024{\natexlab{a}})Zou, Cheng, Cao, Huang, Shan, and Zhang]{DBLP:conf/aaai/Zou0CHSZ24}
Zou, Z., Cheng, W., Cao, Y., Huang, S., Shan, Y., and Zhang, S.
\newblock Sparse3d: Distilling multiview-consistent diffusion for object reconstruction from sparse views.
\newblock In \emph{{AAAI}}, pp.\  7900--7908, 2024{\natexlab{a}}.

\bibitem[Zou et~al.(2024{\natexlab{b}})Zou, Yu, Guo, Li, Liang, Cao, and Zhang]{zou2024triplane}
Zou, Z.-X., Yu, Z., Guo, Y.-C., Li, Y., Liang, D., Cao, Y.-P., and Zhang, S.-H.
\newblock Triplane meets gaussian splatting: Fast and generalizable single-view 3d reconstruction with transformers.
\newblock In \emph{Proceedings of the IEEE/CVF Conference on Computer Vision and Pattern Recognition}, pp.\  10324--10335, 2024{\natexlab{b}}.

\end{thebibliography}
\bibliographystyle{icml2025}




\end{document}